\documentclass{article}

\usepackage{iclr2026_conference,times}

\usepackage{xcolor}
\usepackage{graphicx}
\usepackage{subcaption}
\usepackage{algorithm}
\usepackage{algpseudocode}
\usepackage[most]{tcolorbox}
\usepackage{colortbl}
\usepackage{booktabs}
\usepackage{tabularx}
\usepackage{wrapfig}
\usepackage[inline]{enumitem}
\usepackage{amsmath}

\usepackage{placeins}
\makeatletter
\def\onedot{\futurelet\@let@token\@onedot}
\def\@onedot{\ifx\@let@token.\else.\null\fi\xspace}
\makeatother
\usepackage{xspace}
\def\eg{\emph{e.g}\onedot}
\def\ie{\emph{i.e}\onedot}

\newcommand{\model}[1]{\textsc{FoveAgent-LSTM}}
\newcommand{\qwen}[1]{\textsc{Qwen2.5-VL-3B-Instruct}}
\newcommand{\foveaqwen}[1]{\textsc{FoveAgent-Qwen}}

\newcommand{\myparagraph}[1]{\vspace{3pt}\noindent\textbf{#1}}

\definecolor{taskVisualParityColor}{HTML}{D55E00}
\definecolor{taskStateMachineColor}{HTML}{0072B2}
\definecolor{taskRecallColor}{HTML}{009E73}
\definecolor{taskFindingZerosColor}{HTML}{CC79A7}

  \newcommand{\taskVisualParity}{\textcolor{taskVisualParityColor}{\textsc{Visual Parity}}}
  \newcommand{\taskStateMachine}{\textcolor{taskStateMachineColor}{\textsc{State Machine}}}
  \newcommand{\taskRecall}{\textcolor{taskRecallColor}{\textsc{Recall}}}
  \newcommand{\taskFindingRoots}{\textcolor{taskFindingZerosColor}{\textsc{Finding Roots}}}

\usepackage{url}
\definecolor{linkblue}{HTML}{1A5FB4}
\usepackage[colorlinks=true,
            linkcolor=linkblue,
            citecolor=linkblue,
            urlcolor=linkblue,
            backref=page]{hyperref}
\usepackage[capitalize]{cleveref}
\crefname{section}{Sec.}{Secs.}
\Crefname{section}{Sec.}{Secs.}
\crefname{table}{Tab.}{Tabs.}
\Crefname{table}{Tab.}{Tabs.}

\usepackage{orcidlink}

\iclrfinalcopy

\begin{document}

\title{On Locality and Length Generalization \\
in Visual Reasoning}

\author{Pulkit Madan$^\dagger$ \orcidlink{0009-0007-8304-1041} \quad
\textbf{Sanjay Haresh}$^\dagger$ \orcidlink{0009-0000-5096-3217} \quad
\textbf{Reza Ebrahimi} \orcidlink{0000-0002-4925-3477} \quad
\textbf{Sunny Panchal} \orcidlink{0009-0009-8864-2422} \\
\textbf{Apratim Bhattacharyya} \orcidlink{0000-0002-3481-9546} \quad
\textbf{Roland Memisevic} \orcidlink{0009-0009-4168-9397} \\
Qualcomm AI Research\thanks{Qualcomm AI Research is an initiative of Qualcomm Technologies, Inc.\\$\dagger$ Equal contribution.}\\
\texttt{\{pmadan, sanjayh\}@qti.qualcomm.com}\\
}

\maketitle

\begin{abstract}
A striking feature of the human visual system is
that it ingests visual information through a series of local foveated glimpses, rather than a single global computation.
This makes human vision distinctly different from most popular
computer vision models in use today, which input images globally and
in a single shot.
A natural question therefore is whether local, sequential vision models
may provide any fundamental computational benefits  in addition to
being biologically more plausible than global models.
In this work, we investigate this question
from the perspective of visual state tracking and length generalization.
Inspired by recent studies of length generalization in language models,
we study the behavior of vision models trained on simple vision tasks that
require the aggregation of local information across an image.
Our experiments reveal that, similar to language models, vision models
can learn to exploit global shortcuts and thereby fail to generalize
over task length or complexity.
We also show that recurrent vision policies based on strictly
local perception can mitigate these failures, thereby allowing models to
generalize on these tasks.
Our results show that local attention may be an essential overlooked requirement for robust compositional generalization. 
\end{abstract}

\section{Introduction}

Current state-of-the-art vision models have shown human level performance on tasks such as image captioning and visual question answering~\citep{Qwen2.5-VL,gpt5p4,claudesonnet4p6}.
This success is built on models which ingest an image in a single forward pass to create an encoding of the global contents of the image. 
\textit{E.g.}, transformer based models encode images in a sequence of tokens and (self-)attend to these tokens at every step of a reasoning process.
This mechanism is different from the way humans process images,
which is based on local glimpses connected through saccades (\eg, \citep{hayhoe2005eye}).
This raises the question of whether this type of sequential processing
is purely an evolutionary artifact or if it is beneficial,
or even necessary, for human-level multimodal intelligence.

\begin{figure}[t]
    \centering
    \includegraphics[width=\textwidth]{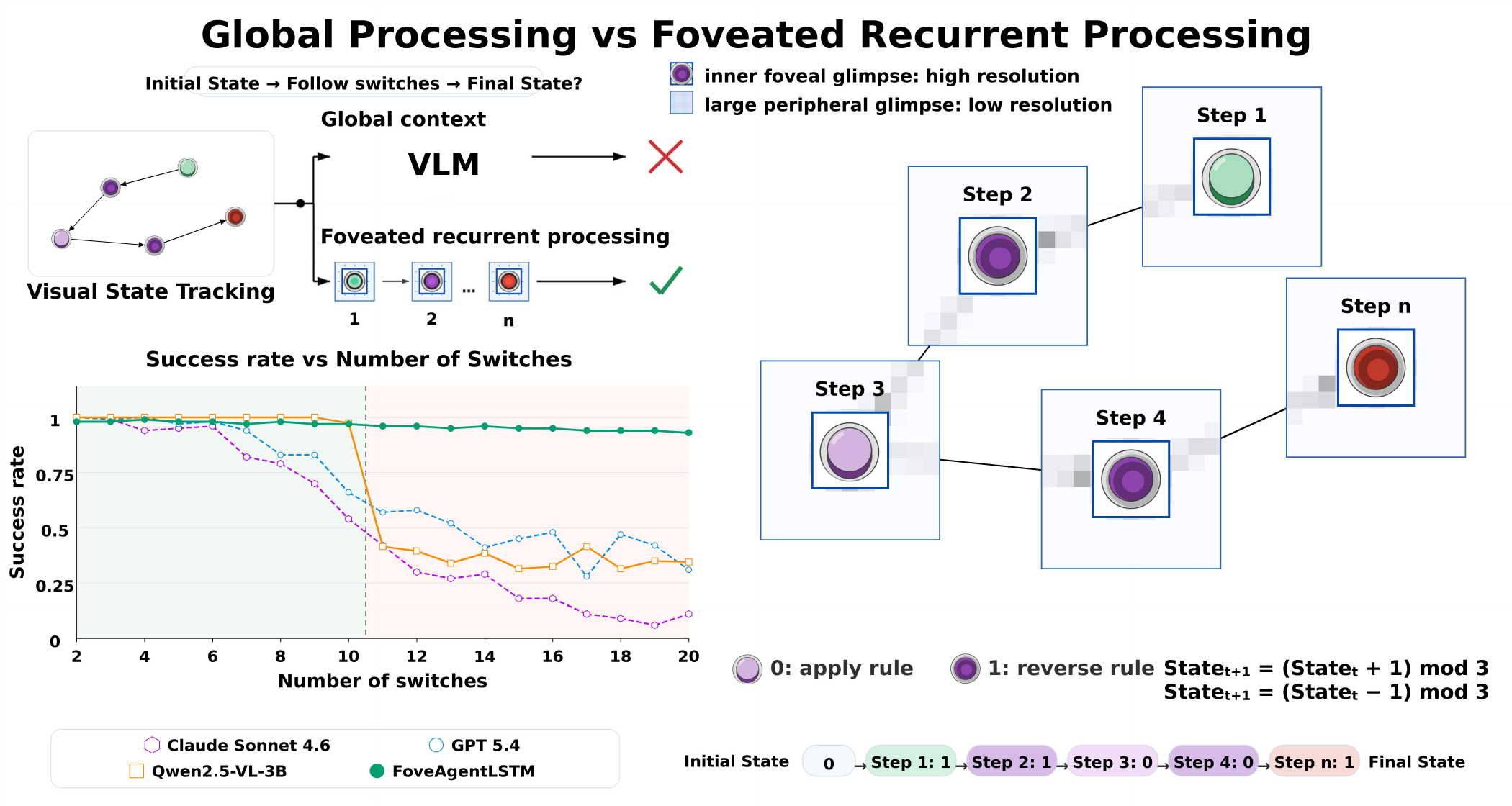}
    \vspace{-0.75cm}
    \caption{Overview of our work's central question: How should visual reasoning models process spatially distributed evidence when test-time task length exceeds the training range?
  Global models, that process the full image in a single pass, can learn shortcuts that fail 
  out-of-distribution. 
  Foveated recurrent processing, in contrast, decomposes the task into repeated 
  local observations and state updates, 
  which supports generalization to longer visual sequences.}
    \label{fig:teaser}
    \vspace{-0.4cm}
\end{figure}

To shed light onto this question, we introduce a set of simple visual reasoning tasks that humans would solve by following a
trajectory of local glimpses over the image.
The  problems involve aggregating local 2D information to derive the
state of a system represented diagrammatically in the image (\cref{fig:teaser}).
The complexity of these visual reasoning problems can be defined by their \emph{length}, which measures the  minimum number of steps
required to solve them.
The main challenge we consider in this work is that of extrapolation \textemdash{} solving longer problems without explicit training \textemdash{} usually referred to as \emph{length generalization}.
Extrapolation requires models to go beyond memorization and towards true
compositional understanding,
requiring strategies that are independent of the length of problem.
Following a trajectory of local glimpses is an example of a strategy
that is independent of problem length.

Our visual reasoning problems are inspired by standard tasks widely used to study
length generalization in language models, such as the task of determining the parity
of a binary sequence.
Recent work has shown that transformer language models fail to length-generalize in such tasks
\citep{dubois2020location,anil_lengthgeneralization,faithandfate,NEURIPS2024_3107e4bd,deletang2023neural,ebrahimi2024your,zhou2024algorithms}.
This failure is widely attributed to the fact that these models learn global ``shortcut'' solutions \citep{liu2023transformers, li2025how},
which fail to represent the compositional
structure in these tasks and therefore do not generalize
out-of-distribution (OOD).
Previous studies \citep{NEURIPS2024_3107e4bd,ebrahimi2024your} trace these shortcuts to the global attention mechanism: transformers struggle to retrieve the right information in the right order from their
context window, and thus fail to sequentially retrieve a required
token at the right time.

While the inability to length-generalize in such sequential tasks has also been shown to hold
for state-space models (SSM) \citep{merrill2024illusion, sarrof2024the,cirone2024theoretical,shakerinava2026the,ebrahimi2025revisiting},
nonlinear recurrent networks (RNNs) do not suffer from this shortcoming \citep{giles1992learning,casey1996dynamics,deletang2023neural}. This has been argued to be due to their ability to decompose a reasoning task inductively into a genuine step-by-step computation \citep{ebrahimi2026on}.%
In this work, we show a similar result in visual reasoning.
However, in contrast to the existing work, we also show that recurrence is only one of two
conditions that enables OOD generalization.
The other is locality of perception, that is, the ingestion of information through a sequence
of local glimpses.
In fact, we show that a recurrent network exposed to a global view of
the image does \emph{not} learn to generalize OOD.

Our work thus elucidates the importance of local perception. 
This is an aspect of reasoning that is naturally obscured in textual versions of these tasks.  
The reason is that in textual tasks, 
the relevant information is already decomposed into meaningful symbols
and passed as tokens to the model sequentially.
In contrast, the tasks we present in our work require a model not only to follow an abstract
reasoning process but also to gather the local information required to complete the task.
Our results suggest that the task of information gathering is an important
overlooked aspect of general reasoning tasks. And it suggests local perception to be 
a key ingredient enabling a model to gather information in scenarios unseen during training.
In line with previous findings in language modeling, we further show that 
generalization on state tracking tasks is distinctly different from generalization on recall tasks, and that local perception is \emph{not} required (nor beneficial) for the latter.

  Our main contributions in detail are:
  \begin{enumerate}
      \item We propose length generalization as a fundamental testbed for visual reasoning, requiring models to extrapolate
  to images with more objects, longer visual dependencies, and greater task complexity than seen during training.

      \item We show that current vision and vision-language models, despite strong in-distribution performance, fail to
  extrapolate in this setting, exposing a gap in visual reasoning that is not captured by standard benchmark evaluations.

      \item We trace this failure to global perception: when models process the entire image at once, they can learn to use perceptual shortcuts that break under increased visual complexity.

      \item We demonstrate that recurrent and strictly local perception enables length generalization. 
      We also show that size and resolution of local glimpses is crucial for robust generalization.

      \item Across three synthetic tasks and one real-world task, our results suggest that achieving flexible, human-like reasoning may require a return to biologically inspired, sequential attention policies.
  \end{enumerate}


\section{Test-bed for Length Generalization}
\label{section:roleoflocalperception}

In the following, we first present the tasks designed to evaluate generalization
in visual reasoning tasks and subsequently we present the datasets used for each task.

\subsection{Length Generalization Tasks}
\label{sec:tasks}

\begin{figure*}[t]
    \centering
    \begin{subfigure}[b]{0.45\textwidth}
        \centering
        \fbox{\includegraphics[width=\linewidth,keepaspectratio]{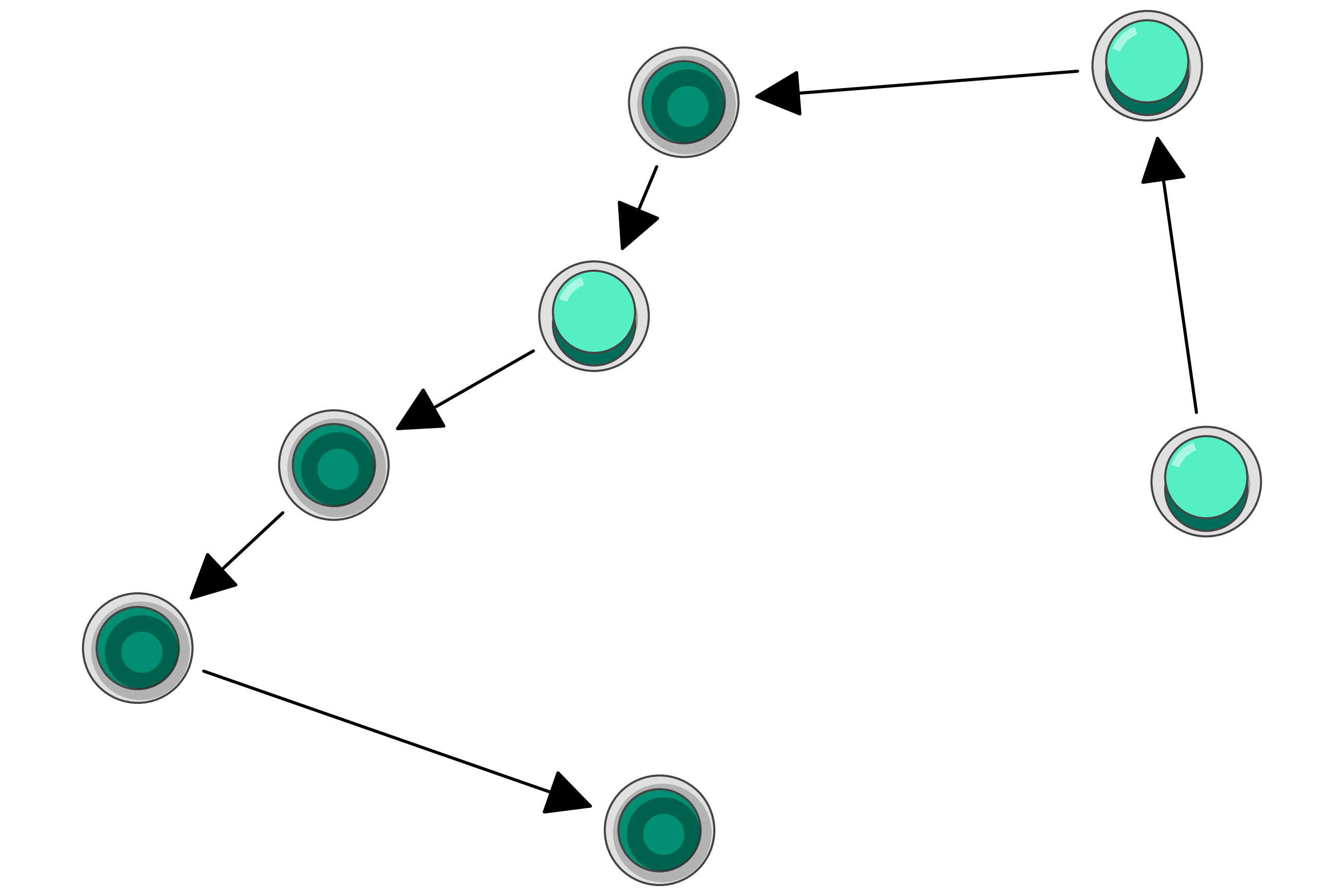}}
        \caption{\taskVisualParity{}}
        \label{fig:task_switch_toggle}
    \end{subfigure}
    \hspace{0.02\textwidth}
    \begin{subfigure}[b]{0.45\textwidth}
        \centering
        \fbox{\includegraphics[width=\linewidth,keepaspectratio]{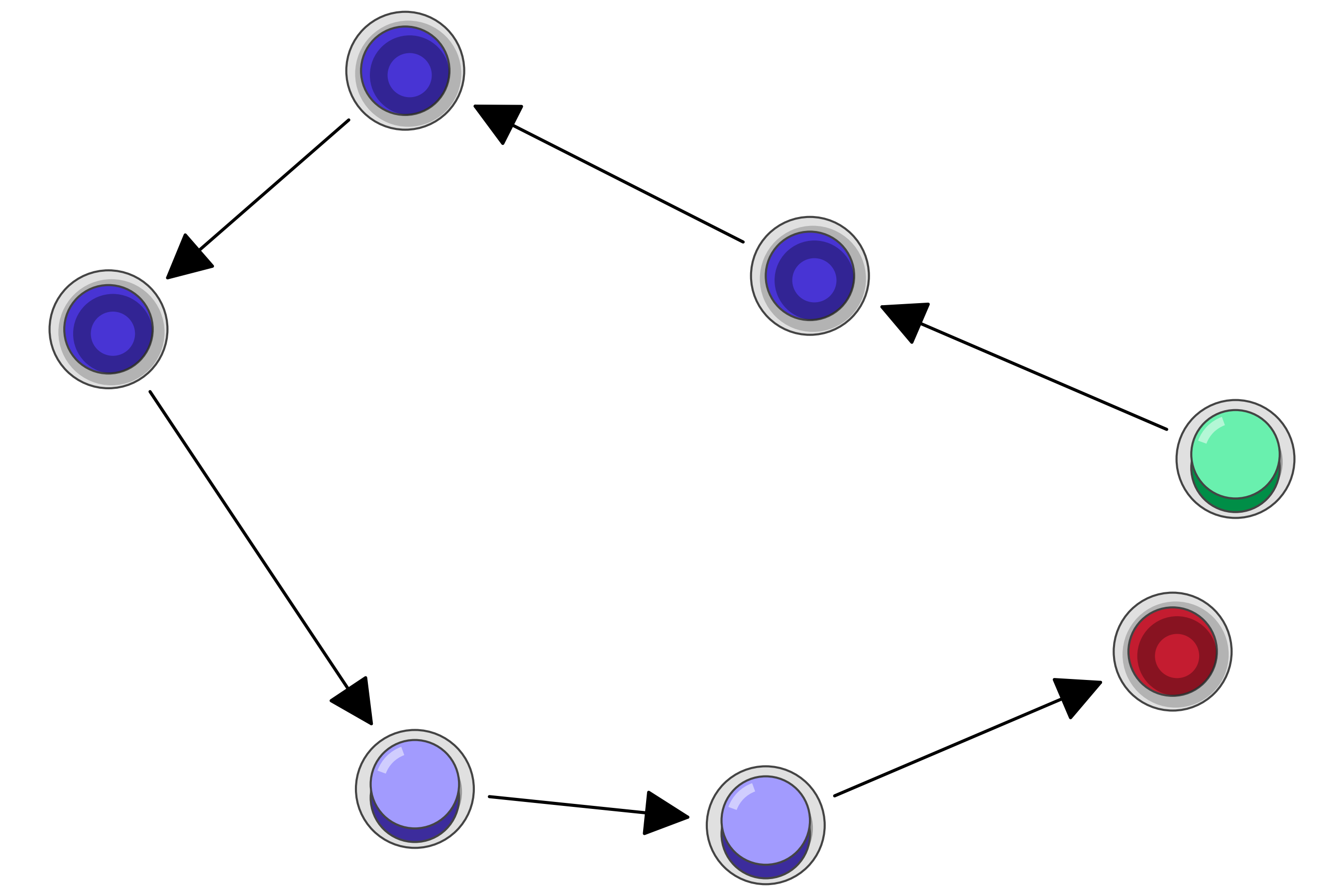}}
        \caption{\taskStateMachine{}}
        \label{fig:task_state_machine}
    \end{subfigure}

    \vspace{0.3cm}

    \begin{subfigure}[b]{0.45\textwidth}
        \centering
        \fbox{\includegraphics[width=\linewidth,keepaspectratio]{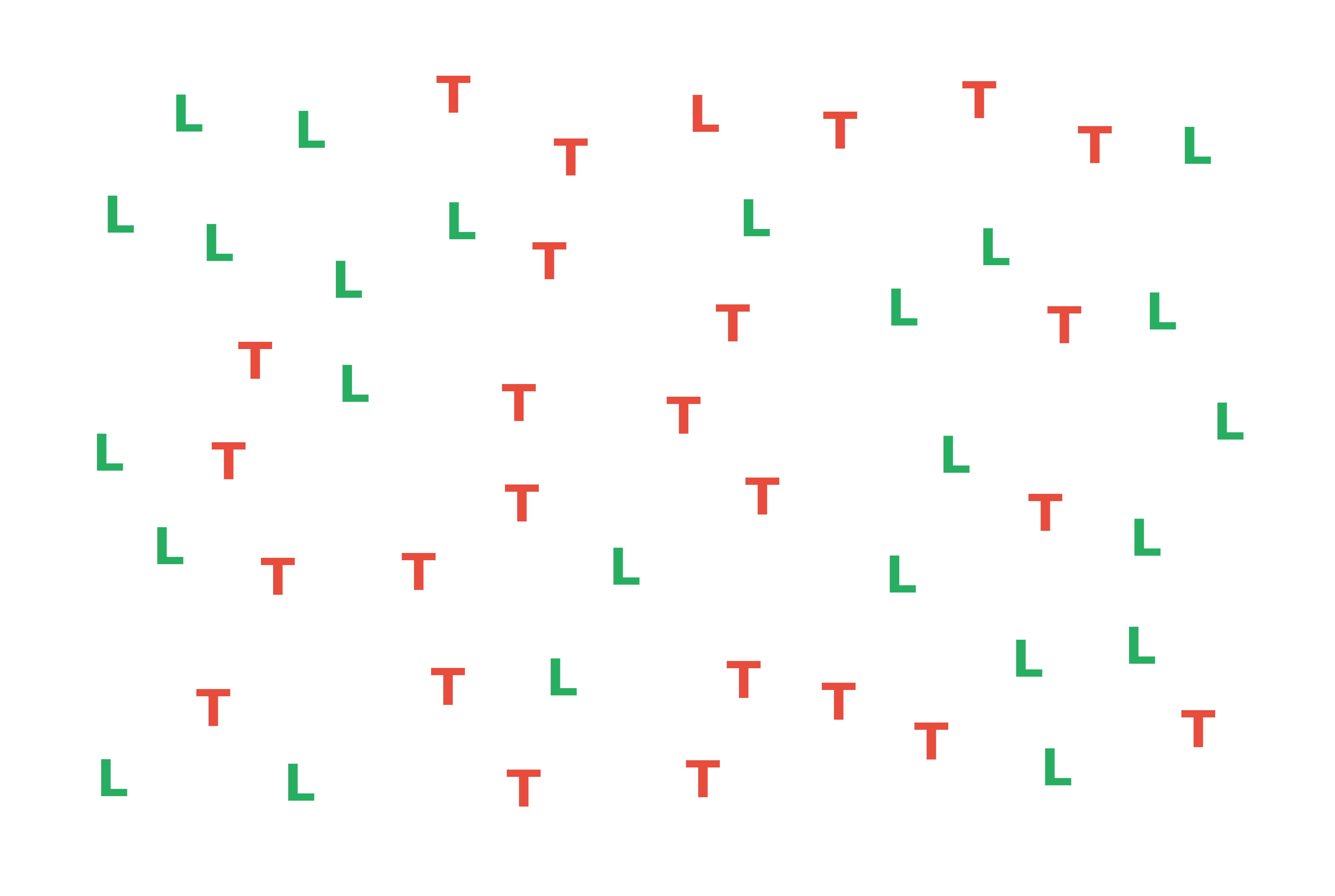}}
        \caption{\taskRecall{}}
        \label{fig:task_conjunctive_search}
    \end{subfigure}
    \hspace{0.02\textwidth}
    \begin{subfigure}[b]{0.45\textwidth}
        \centering
        \fbox{\includegraphics[width=\linewidth,keepaspectratio]{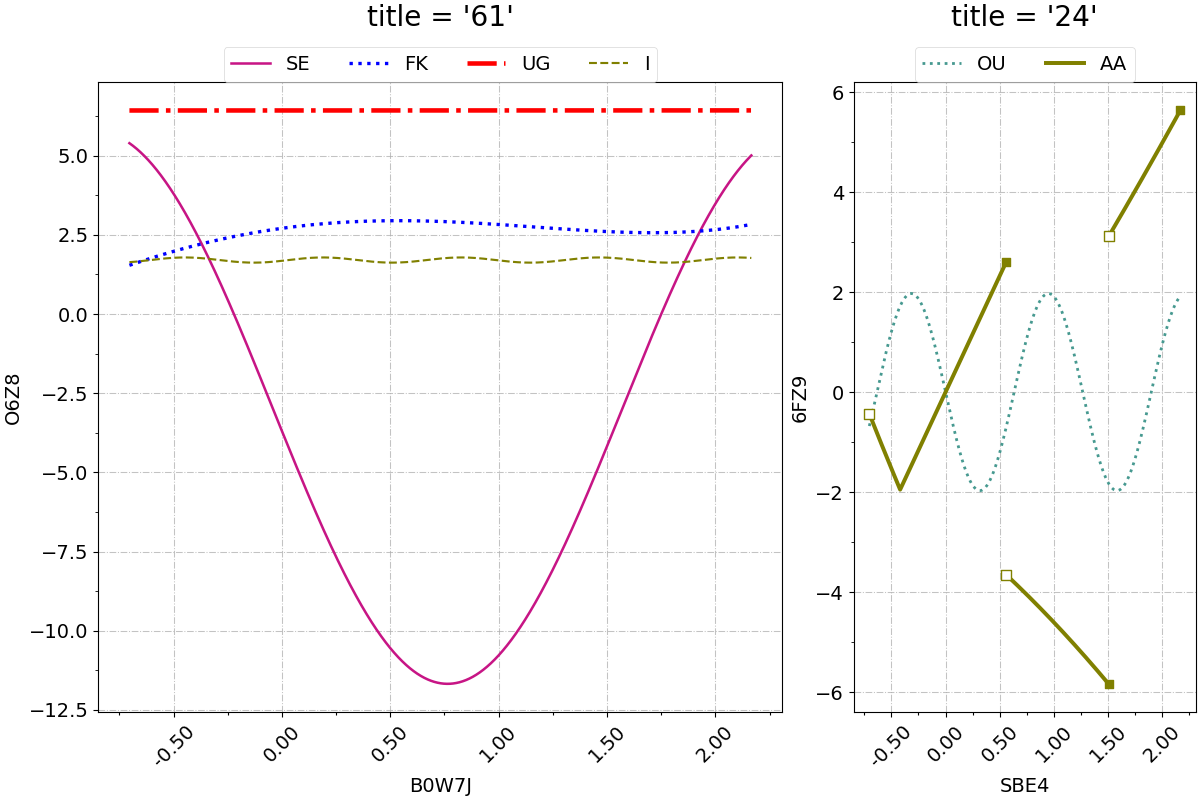}}
        \caption{\taskFindingRoots{}}
        \label{fig:task_plottwist}
    \end{subfigure}

    \vspace{-0.2cm}

    \caption{
  Examples from our visual reasoning test-bed for length generalization.
  \textbf{Top:} \taskVisualParity{} and \taskStateMachine{} require aggregating switch states over a variable-length visual sequence, with \taskStateMachine{} additionally requiring order-sensitive state updates.
  \textbf{Bottom-left:} \taskRecall{} requires visual retrieval in the presence of clutter (no state tracking).
  \textbf{Bottom-right:} \taskFindingRoots{} is a real-world plot-reasoning task requiring localization of relevant subplots, identification of specific functions, and reasoning over its roots.
  }
    \label{fig:tasks_all}
      \vspace{-0.5cm}
\end{figure*}

We consider visual reasoning problems 
that require a model to determine the state of an underlying system 
encoded as local information distributed over the image. 
We control the task length along the two axes: \emph{resolution}, i.e. {the spatial resolution of the image},  and \emph{task length},
i.e., the number local pieces of information encoding the state. 
We note that these two dimensions are separate but can interact with one another, 
because for a fixed
task length, the density of entities is dependent on the resolution.
\cref{fig:tasks_all} shows examples of the tasks described in detail below.

\myparagraph{\taskVisualParity{}.} This task serves as a visual analog to the classic bit-string Parity problem \citep{dubois2020location,anil_lengthgeneralization,faithandfate,NEURIPS2024_3107e4bd}. The input consists of
N binary ``switches'' (toggleable  states: \textit{pressed / $1$} or \textit{not-pressed / $0$}) placed at random positions across the image. The objective is to compute the global parity of the system, or equivalently 
determining if the sum of \textit{pressed} states is odd or even.

Crucially, this task is subtly different from the customary Parity task used to study length generalization in language models: In the textual domain, an autoregressive model ingests a pre-ordered, linear sequence of bits. In our visual setting, the bits are not fed sequentially. Instead, the relevant information is spatially distributed. Consequently, the model must locate the switches within the 2D canvas and 
aggregate all of their states to determine the global state of the system. 
An example of the task can be seen in \cref{fig:task_switch_toggle}.

\myparagraph{\taskStateMachine{}.} While Visual Parity requires the model to aggregate information across local positions on the image, the task is inherently permutation-invariant,
that is, the order of information aggregation does not matter.
To rule out permutation-invariance as a confounding factor, we also construct a task
in which the order is relevant.
To this end, following \citet{liu2023transformers},
we construct a task that simulates a state machine realizing the action of the dihedral group:
Starting from a pre-defined state,
a \textit{non-pressed / $0$} switch applies an update rule 
to the current state.
States are numerical, and initial the update rule is \textit{increment}.
A \textit{pressed / $1$} switch reverses the current update rule, 
such that the \textit{non-pressed} switch henceforth \textit{decrements} the state.
This will be the case until another \textit{pressed} switch is encountered, 
which changes the meaning back to \textit{increment}, and so on.
To denote the order, we render arrows between the switches. 
The starting switch is denoted using the color green, the ending switch is denoted using the color red, and the intermediate switches are denoted using the color purple. 
All increment and decrement operations are modulo $3$,
so that at inference time, the model never encounter values it has never seen during training. \cref{fig:task_state_machine} shows an example of the task; a formal definition of this construction is provided in
\cref{appsec:dataset_details}.

\myparagraph{\taskRecall{}.}
    Out-of-distribution generalization in the tasks above demonstrates a model's ability to
    perform \emph{state tracking}.
    It is important to note that, while transformer-based language models
    cannot length-generalize in these tasks, they can length-generalize in tasks
    involving key-value retrieval (or ``recall''). 
    These are tasks in which a model needs to determine the value associated 
    with a given key after seeing a sequence of
    interleaved key-value pairs (\eg \citep{phan2025delayed}).

    We therefore include a control setting, in which state tracking is not required to successfully
    generalize out-of-distribution.
    Following \citet{campbell2024understanding}, we define the task as a search problem, 
    in which the model needs to identify if a specific target object is present (True or False) 
    among a set of distractors with multiple overlapping attributes.
    Specifically, consider the task of finding a red letter \textcolor{red}{``L''} 
    within a visual scene showing red letters \textcolor{red}{``T''} 
    and green letters \textcolor{green}{``L''}. 
    Neither color nor shape alone is sufficient to identify the target object. 
    To test generalization, we introduce out-of-distribution cases by increasing the number
    of objects beyond the number shown during training, which adds visual clutter and complexity. 
    To reduce overlap, we increase the size of the canvas jointly with the number of distractors.
    This setup evaluates both recall and robustness under distribution shift.
    Successful generalization is non-trivial, 
    as it requires a form of feature binding \citep{campbell2024understanding}. 
    However, it does
    \emph{not} require state tracking and should solvable using a global, parallel 
    computation over the image similar to object detection.
    An example of the task can be seen in \cref{fig:task_conjunctive_search}.

\myparagraph{\taskFindingRoots{}.}
We also include a task based on mathematical reasoning over plots to evaluate performance
on a real-world task that can be solved naturally with the help of state tracking. 
Understanding of mathematical plots is an important real-world task, for which 
data can nevertheless be generated synthetically. 
This makes it possible to control the amount of local visual information that needs 
to be aggregated in both in-distribution and out-of-distribution settings.
We use the \textit{Finding-Zeros} task from the MathSearch benchmark \citep{madhan2024mathsearch},
where the goal is to determine the zeros of a function from a plot of the function 
in the presence of distractors, such as other functions and subplots.
An example of the task can be seen in \cref{fig:task_plottwist}.
A natural way to answer a question, such as 
\textit{``Find the roots of function SE in subplot 61"}, 
is to use iterative visual search to first localize the relevant subplot, 
to subsequently locate the target function and coordinate axes, 
and to finally pinpoint the intersection of the curve with the horizontal axis ($y=0$).

To generate out-of-distribution scenarios, we set the number of zeros and/or the number 
of subplots at test time to different values than during training.

\subsection{Datasets}
\label{sec:data}

We create Gymnasium environments~\citep{towers2024gymnasium} for the tasks 
in \cref{sec:tasks} along with oracle policies. 
Using these oracle policies, we generate a dataset of trajectories, each containing
a sequence of local glimpses $G_l$, which are small local crops from the larger image, 
corresponding peripheral glimpses $G_p$, which are larger crops centered at the 
same locations,  
delta actions $A$, which are defined as the difference between the current 
glimpse position and the next desired position, 
probes $P$, which are used as extra sub-task supervision as we shall discuss below, 
a task question $Q$, and a task outcome $O$. 
For robustness, we also add noise to the training trajectories by randomly perturbing 
the locations of glimpses.

For each task, we create in-distribution (InD) and out-of-distribution (OOD) settings by varying image resolution and task complexity.
For all our experiments, we only use the InD dataset for training and run evaluations on both InD and OOD settings. Unless specified otherwise, we use $1200\times800$ image 
resolution as the InD resolution. 
For \taskVisualParity{} and \taskStateMachine{}, 
we use $2-10$ switches in the InD setting 
and $10-20$ switches in the OOD setting.
For \taskRecall{}, 
we use $<10$ distractors in the InD setting and $>10$ distractors in the OOD setting. 
Finally, for \taskFindingRoots{},
we use $1$--$6$ roots and $1$--$6$ subplots in the InD setting,  
and we consider two separate OOD settings: we set the number of roots to 
$7$--$10$ for one (setting ``OOD-numroots'') and 
the number of subplots to $7$--$9$ for the other (setting ``OOD-subplots)''. 
Additional dataset details are provided in \cref{appsec:dataset_details}.

\begin{figure*}[!t]
    \centering
    \includegraphics[width=0.95\linewidth,trim=1cm 0.65cm 1cm 0.cm]{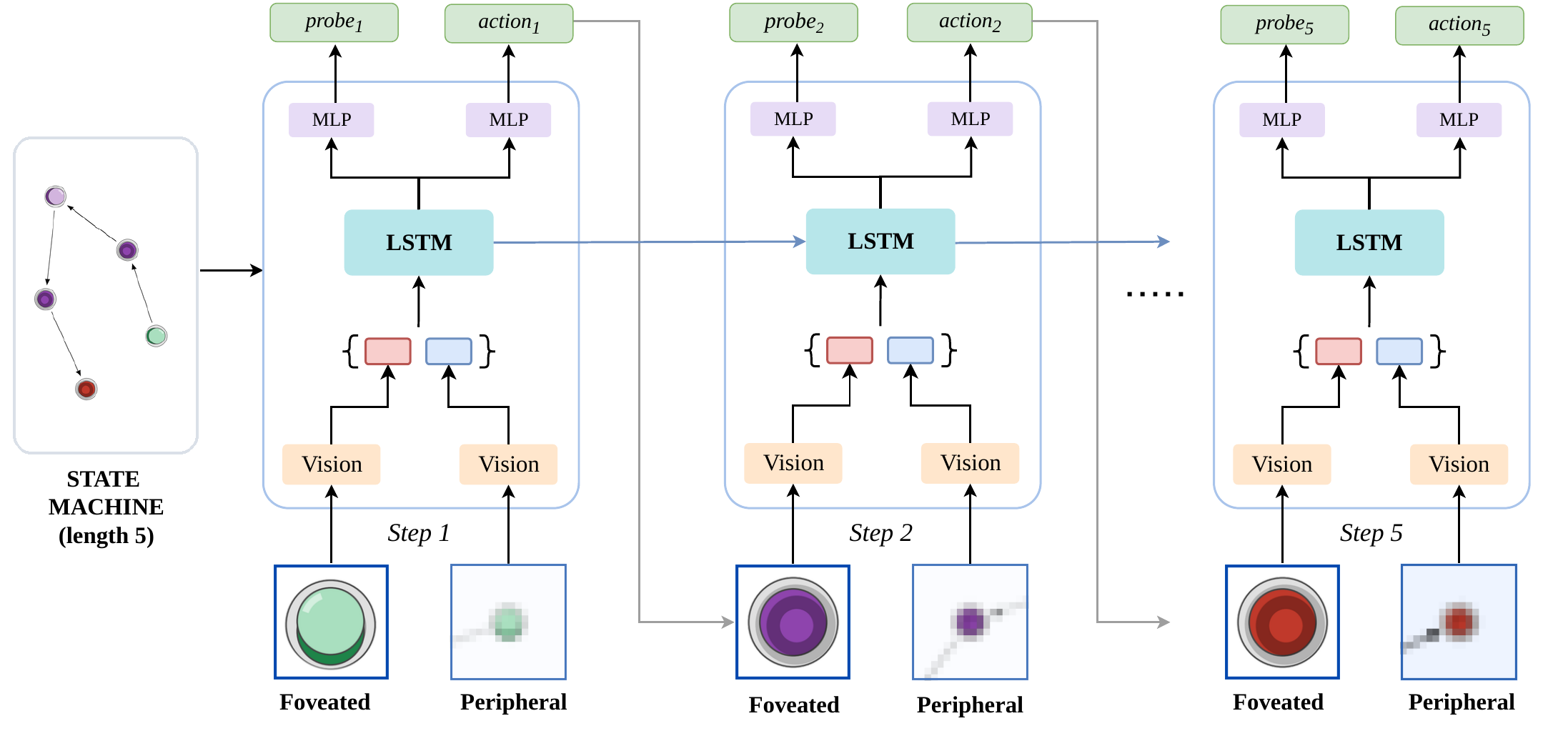}
    \caption{
  Overview of \model{}.
  At each step, the model receives a high-resolution local glimpse together with a low-resolution peripheral glimpse, updates its recurrent state, and predicts the next action and any other task-relevant outputs.
  This step-by-step local processing is different from most standard vision and vision-language models, which process the full image globally in a single forward pass.
  }
    \label{fig:local_model}
\end{figure*}

\section{Recurrent foveation model}
We now describe the recurrent, local vision model we use in our experiments. 
The model ingests visual input through a sequence of foveated and peripheral glimpses, 
similar to earlier models of foveated perception
\citep{NIPS2010_677e0972,ba2014multiple,mnih2014recurrent,Gregor2015DRAWAN}.
We illustrate the architecture, which we refer to as \model{}, in \cref{fig:local_model}.

We hypothesize that the local glimpse can facilitate mainly the processing of fine-grained
image details, whereas the peripheral glimpse can facilitate ``navigation'', that is, 
determining positions to move to. 
We shall present evidence for this separation in Section~\ref{sec:experiments}.


\myparagraph{Overview and Step-by-step Operation.}
For any given vision task, 
the model starts from a fixed initial location, which is provided as an input to the model.
The model then iteratively processes the information encoded in the local glimpses and
generates displacements to determine the next glimpse location as well as any 
additional task-specific information. 
The model continues this iterative process 
until it has acquired the required information to solve the task, 
at which point it outputs the solution.

\myparagraph{Architecture.} The \model{} model consists of a recurrent LSTM backbone as shown in \cref{fig:local_model}. 
The foveated and the peripheral glimpses are each encoded by a 
ResNet~\citep{he2016deep} encoder, 
which output features are concatenated and passed to the LSTM backbone.
Both glimpses are square in shape and the size of the peripheral glimpse $\text{G}_{\text{p}}$ is $4$ $\times$ the size of the foveated glimpse $\text{G}_{\text{l}}$. 
However, both glimpses are resized to the same fixed resolution, 
which we refer to as the \textit{sensor resolution}. 
We use a fixed sensor resolution to mimic the human visual system 
and also to restrict the amount of information in the peripheral glimpse and thus  
to reduce the potential for learning visual ``shortcuts''.

At every timestep, the \model{} outputs the action along with the output of a 
probe (described next).
The action encodes the displacement specifying the next location to be extracted.
Specifically, the displacement is determined by a tuple of the form (angle, magnitude): $(\theta, d)$, where $0 \leq \theta \leq 2\pi$ and  $0 \leq d \leq \text{G}_{\text{p}}$.
The next location, $x_{t+1}$, is defined as $x_{t+1} = x_t + d u_t(\theta)$, 
where $x_t$ is the current position and $u_t(\theta)$ is a unit vector pointing in the direction 
given by the angle $\theta$. 
As $0 \leq d \leq \text{G}_{\text{p}}$, the new local glimpse center $x_{t+1}$ is constrained to be within the peripheral glimpse location.
Additionally, the action also contains a stop bit, $s$, which is 1 when the model decides to output the solution, and 0 otherwise.

\myparagraph{Probes.}
At every time step, in addition to an action, the \model{} can output the value of a probe which makes it possible to 
add supervision signals beyond final answer 
accuracy during training \citep{BhattacharyyaP024}.
In the \taskVisualParity{} and \taskStateMachine{} tasks, 
probes are 0, 1 or null if the state in the current local glimpse is 0, 1 or not visible, respectively.
This kind of information is analogous to aligned chain-of-thought 
supervision used in textual tasks, where it can drastically improve data efficiency for learning \citep{ebrahimi2026on}. 
In the case of \taskFindingRoots{}, the probe encodes whether a function root is 
observed or not.

  \begin{figure*}[!t]
    \centering

    \makebox[\linewidth][c]{%
      \includegraphics[width=0.98\linewidth]{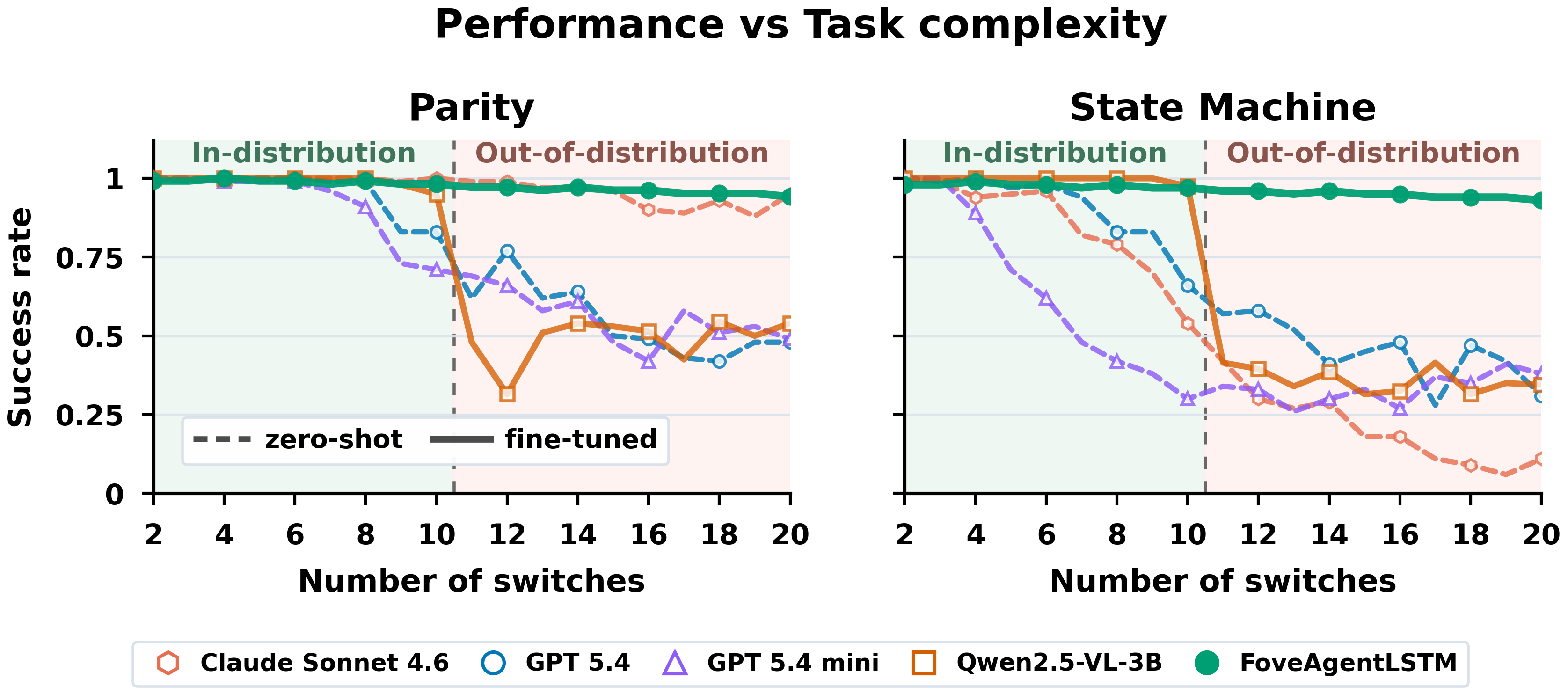}%
    }

    \vspace{0.6em}

    \makebox[\linewidth][c]{%
      \includegraphics[width=0.98\linewidth]{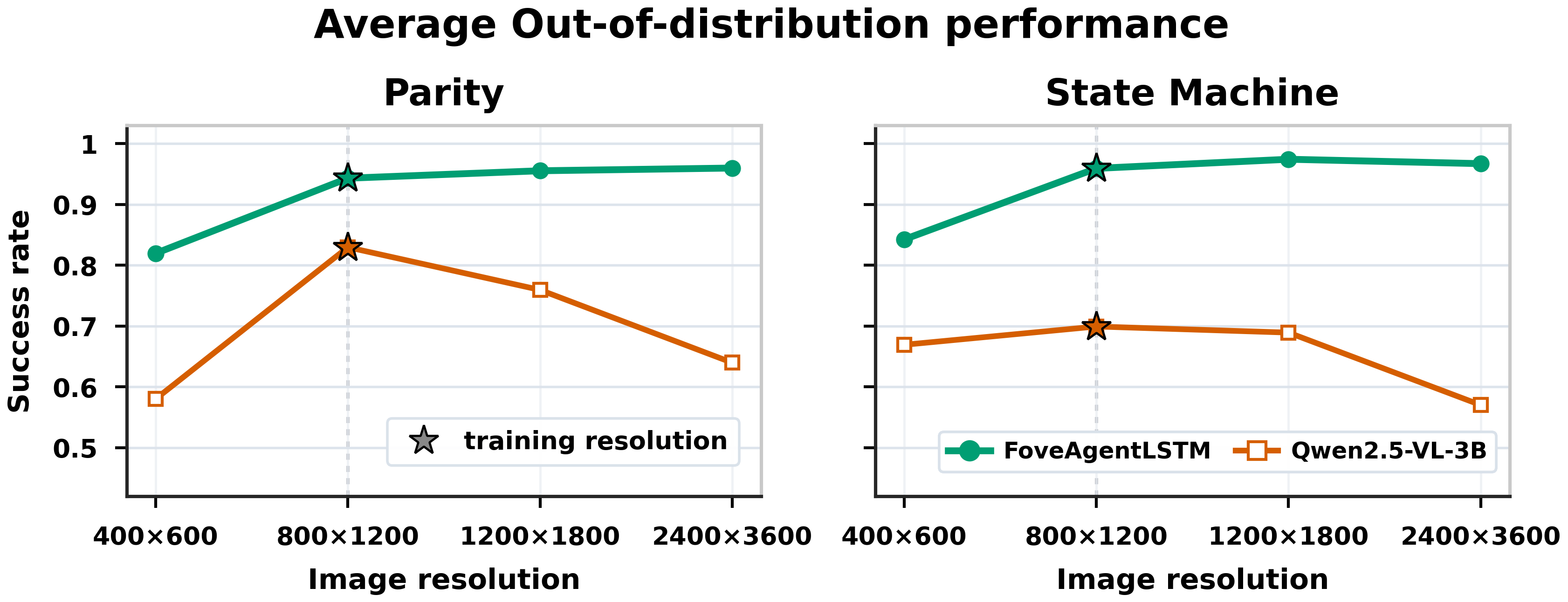}%
    }

 \caption{
  Success rates on \taskVisualParity{} and \taskStateMachine{} as a function of task complexity and image resolution.  
  \textbf{Top:} performance for varying numbers of switches (within and beyond the training range). The shaded regions indicate in-distribution and out-of-distribution task lengths.
  \textbf{Bottom:} average OOD performance across image resolutions, 
  with stars marking the training resolution.
  The foveated model maintains high accuracy out-of-distribution, while the performance of the 
  global-view VLM baselines degrades with increasing task length and resolution.  
  }
    
    \label{fig:main_results}
  \end{figure*}

\section{Experiments}
\label{sec:experiments}
We now present evaluation results that support our core hypotheses regarding 
OOD generalization in visual state tracking. We summarize our hypotheses as follows: 
  \begin{itemize}
      \item[H1.] Local perception is necessary for OOD generalization.
      \item[H2.] Recurrence is necessary for OOD generalization.
      \item[H3.] Together, recurrence and local perception are sufficient for OOD generalization.
  \end{itemize}

\myparagraph{Experimental Setup.} 
We vary both the task complexity and image
resolution to create an OOD set as described in \cref{sec:data}.
In this section, we compare the \model{} against \qwen{}, which is
a broadly used state-of-the-art visual reasoning model.
We train the models on the InD set as described in \cref{sec:data} until convergence.
We also perform zero-shot evaluations on a range of other
closed source state-of-the-art models, using the prompts and chain-of-thought instructions reported in \cref{appsec:zeroshot_prompts}.
\vspace{3pt}

\myparagraph{Evaluation Metrics.} We use final answer accuracy as our main metric for evaluation.
For \model{}, in addition to the final answer, we also evaluate whether the model 
visits all locations containing task-relevant local information, \eg, all switches in \taskVisualParity{}. The full evaluation criteria
are described in \cref{appsec:evaluation_metrics}.

\subsection{Parity and State Machine Tasks}
\label{sec:exp_2d_tasks}
To evaluate the ability of the models to perform state tracking,
we use the \taskVisualParity{} and \taskStateMachine{} tasks described in \cref{sec:tasks}.
Note that the InD dataset used for training contains images showing 2 to 10 switches on a 800x1200 resolution canvas.
The OOD evaluation set contains varying image resolutions and tasks with 11 to 20 switches
We first compare the \model{}
with \qwen{}, which we finetune to make predictions using the full high resolution 
image along with the ``chain-of-thought'' description of the target glimpse positions 
to provide the same supervision information as that of \model{}. 
We also compare \model{} with various SOTA models like GPT 5.4 \citep{gpt5p4} 
and Claude Sonnet 4.6 \citep{claudesonnet4p6} using few-shot prompting.

We show the results for the \taskVisualParity{} and \taskStateMachine{} tasks in \cref{fig:main_results}.
The results show that \model{} generalizes across both axes of task complexity, resolution, and task length.
On the other hand, \qwen{}, while able to achieve high accuracy in-distribution, 
shows a sharp decline in performance as the task complexity increases.


For the few-shot closed-source VLMs, we noticed that the degradation with increasing canvas complexity appears to reflect multiple sources of error. In many cases, the models fail to recover the complete task-relevant structure from the image, for example by missing switches or misinterpreting arrows. In other cases, the visual evidence appears to be available but the state-tracking computation still fails.
This highlights that in spite of the intentional simplicity of the task design, these tasks are non-trivial to solve, and challenge both visual information gathering, and state tracking capabilities of the best available VLMs. 
Overall, the results summarized in Figure~\ref{fig:main_results} provide evidence for hypothesis \textbf{H3}.

\begin{figure}[!t]
\centering

\begin{minipage}[t]{0.49\textwidth}
    \centering
    \begin{minipage}[c][3.8cm][c]{\linewidth}
        \centering
        \includegraphics[height=3.8cm]{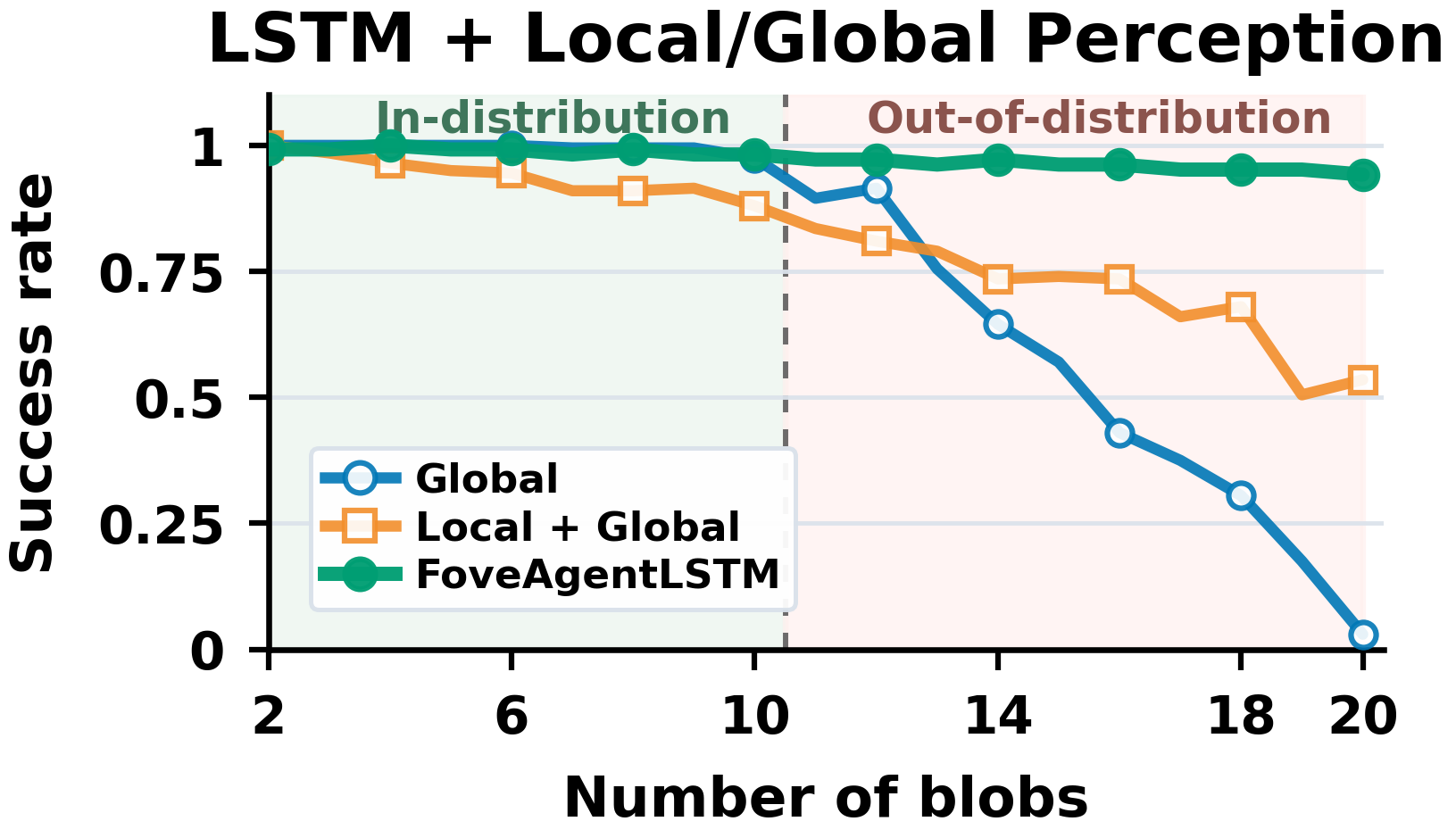}
    \end{minipage}

      \captionof{figure}{
   Effect of the visual interface on length generalization for \taskVisualParity{} with the same recurrent backbone.
  The performance of the \textit{Global} and \textit{Local+Global} variants degrades OOD, while the foveated setup maintains high success using local high-resolution glimpses with low-resolution peripheral context.
  }
    \label{fig:glimpse_design_ablation}
\end{minipage}
\hfill
\begin{minipage}[t]{0.49\textwidth}
    \centering
    \begin{minipage}[c][3.8cm][c]{\linewidth}
        \centering
        \includegraphics[height=3.8cm]{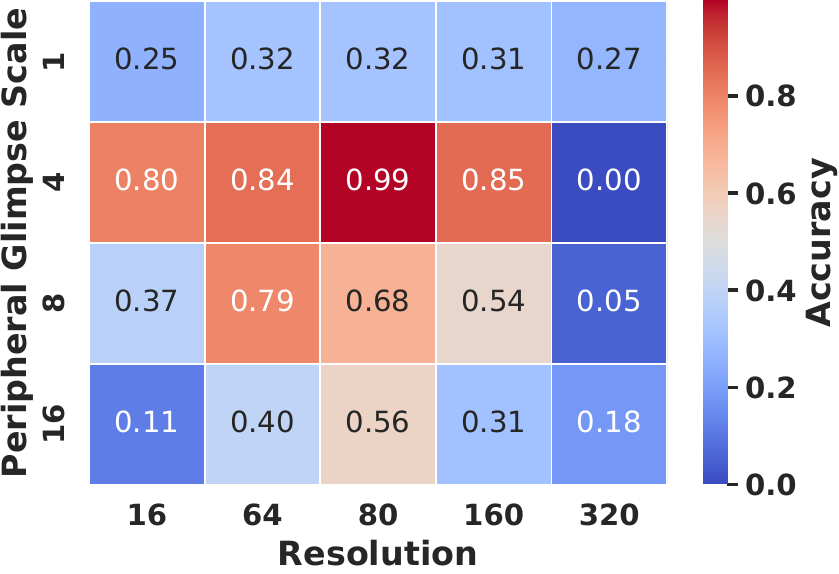}
    \end{minipage}

   \captionof{figure}{
  Effect of peripheral glimpse scale and resolution on \model{}.
  The x-axis varies the resolution of the peripheral glimpse and the y-axis varies the scale of the peripheral glimpse relative to the local glimpse.
  Best performance should require sufficient spatial context for 
  navigation without allowing the peripheral view to be used to compute 
  global shortcuts.
  }
    \label{fig:glimpse_res_size}
\end{minipage}

\end{figure}

\subsection{Local vs Global Perception}
In Section~\ref{sec:exp_2d_tasks}, we show that a model with local visual processing 
and a recurrent backbone is able to generalize OOD in visual state tracking tasks. 
We now examine in more detail which aspects of the model enable OOD generalization. 
To this end, we consider three setups which differ in how visual 
information is provided at each time step:

\begin{enumerate}[label=(\arabic*)]
  \item \textit{Global}: The recurrent model receives the full high-resolution image showing the task 
  \item \textit{Local+Global}: The recurrent model receives local high-resolution crops along with the full high-resolution image showing the task 
  \item \model{}: The model receives local high-resolution crops along with larger peripheral crops at 
  down-scaled resolution (same as in the previous section) but never sees the full image
\end{enumerate} 

Figure~\ref{fig:glimpse_design_ablation} shows that 
although the three settings have similar in-distribution performance,
both \textit{Global} and \textit{Local+Global}, 
in which the model has access to the full high resolution canvas during training, 
show poor OOD generalization. 
The results provide evidence for hypothesis \textbf{H1}.

\subsection{Varying Glimpse Size and Resolution}
\label{sec:ablations}



The use of local receptive fields requires a model to take multiple steps to explore 
the input image and aggregate information before predicting the output.
For most typical tasks, the number of steps required to collect the relevant 
information will be inversely proportional to the glimpse size, as 
larger glimpses cover more canvas area per step.  
This also means that with larger glimpse sizes, 
a recurrent network can process the whole image in fewer steps and 
requires fewer time-steps to back-propagate errors through for learning. 

On the other hand, according to the main thesis of this work, larger glimpses 
also provide an opportunity to learn global shortcuts that do not generalize. 
For example, in the case of Visual Parity, a larger glimpse containing
multiple switches would allow the model to learn a parallel computation 
that does not generalize out-of-distribution. 
This suggests that there is a strict trade-off with respect to 
glimpse size: larger glimpse sizes should benefit learning, and 
smaller glimpse sizes should benefit OOD generalization.

This trade-off also affects the resolution with which the 
glimpse is made available to the model (the sensor-resolution of the glimpse): 
an optimal sensor resolution needs to be high enough to support 
navigating the canvas, but it needs to be low enough to avoid learning 
shortcuts over the glimpse. 
This creates a conflict which foveation can elegantly resolve: 
the high-resolution fovea makes it possible to infer the details relevant 
to the task, while the low-resolution periphery makes it possible to navigate 
the canvas. 

To study the relationship between glimpse size, resolution and resulting
accuracy quantitatively, we vary glimpse sizes and resolution on \taskVisualParity{}.
We show the results using the \model{} model in \cref{fig:glimpse_res_size}.
To jointly capture prediction accuracy and exploration efficiency, we report the metric  $p_{rs} = A / L$ where $r$ denotes the resolution, $s$ denotes the scale of the peripheral glimpse, $A$ denotes the success rate or accuracy and $L$ denotes the average number of steps taken by the model before making a prediction.
We find that there is indeed an optimal resolution and size for 
the peripheral glimpse at resolution 80 and size 4, and increasing 
resolution or size deteriorates performance. 

\begin{figure}[!t]
\centering

\begin{minipage}[t]{0.48\textwidth}
    \centering
    \begin{minipage}[c][3cm][c]{\linewidth}
        \centering
        \includegraphics[height=3cm, width=\linewidth]{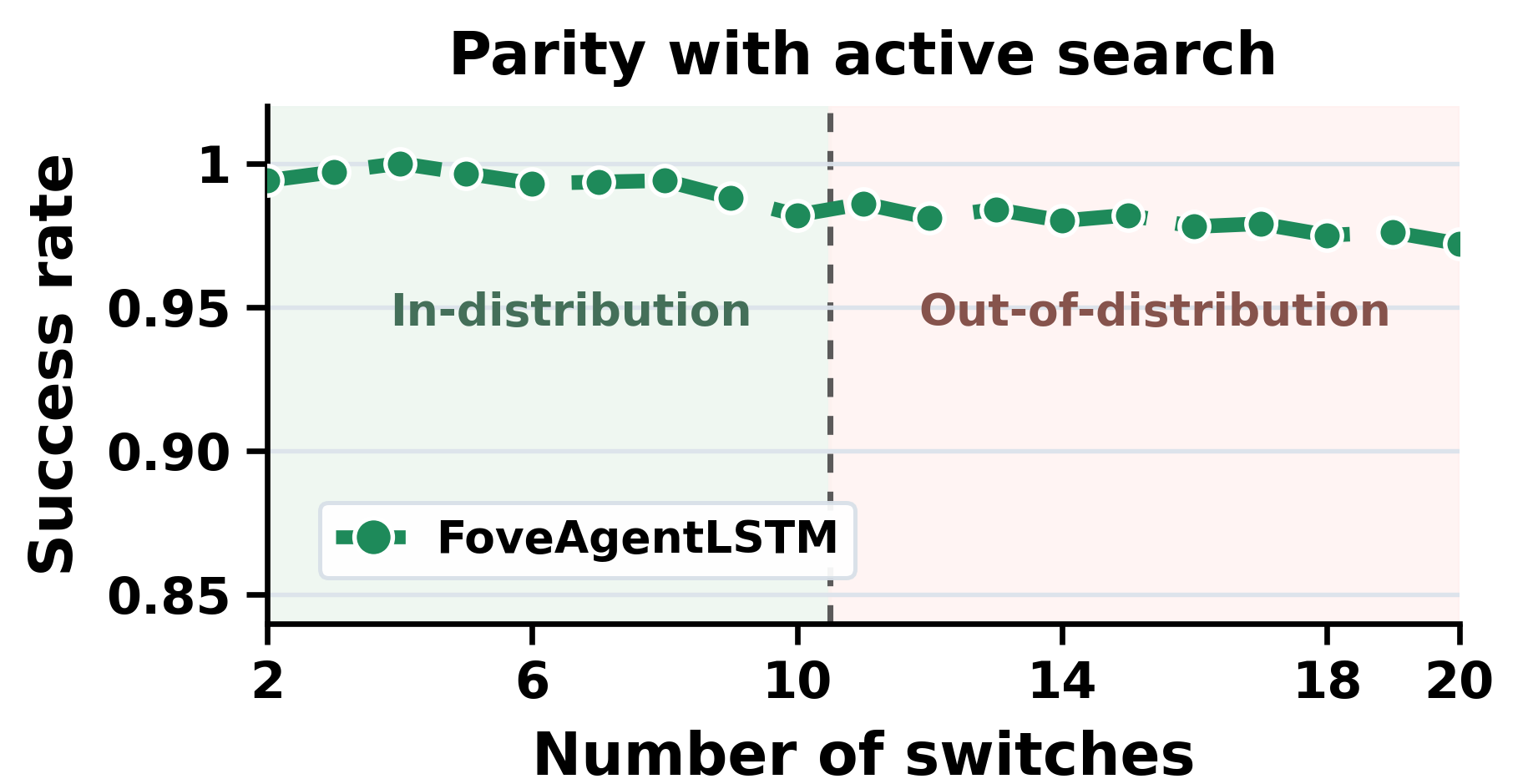}
    \end{minipage}

      \captionof{figure}{
  Generalization on \taskVisualParity{} without traversal arrows.
  \model{} maintains high OOD performance while actively searching for unvisited switches. 
  }
    \label{fig:parity_active_search}
    \vspace{-0.5em}
    
\end{minipage}
\hfill
\begin{minipage}[t]{0.5\textwidth}
    \centering
    \begin{minipage}[c][3cm][c]{\linewidth}
        \centering        
        \includegraphics[height=3cm,width=\linewidth]{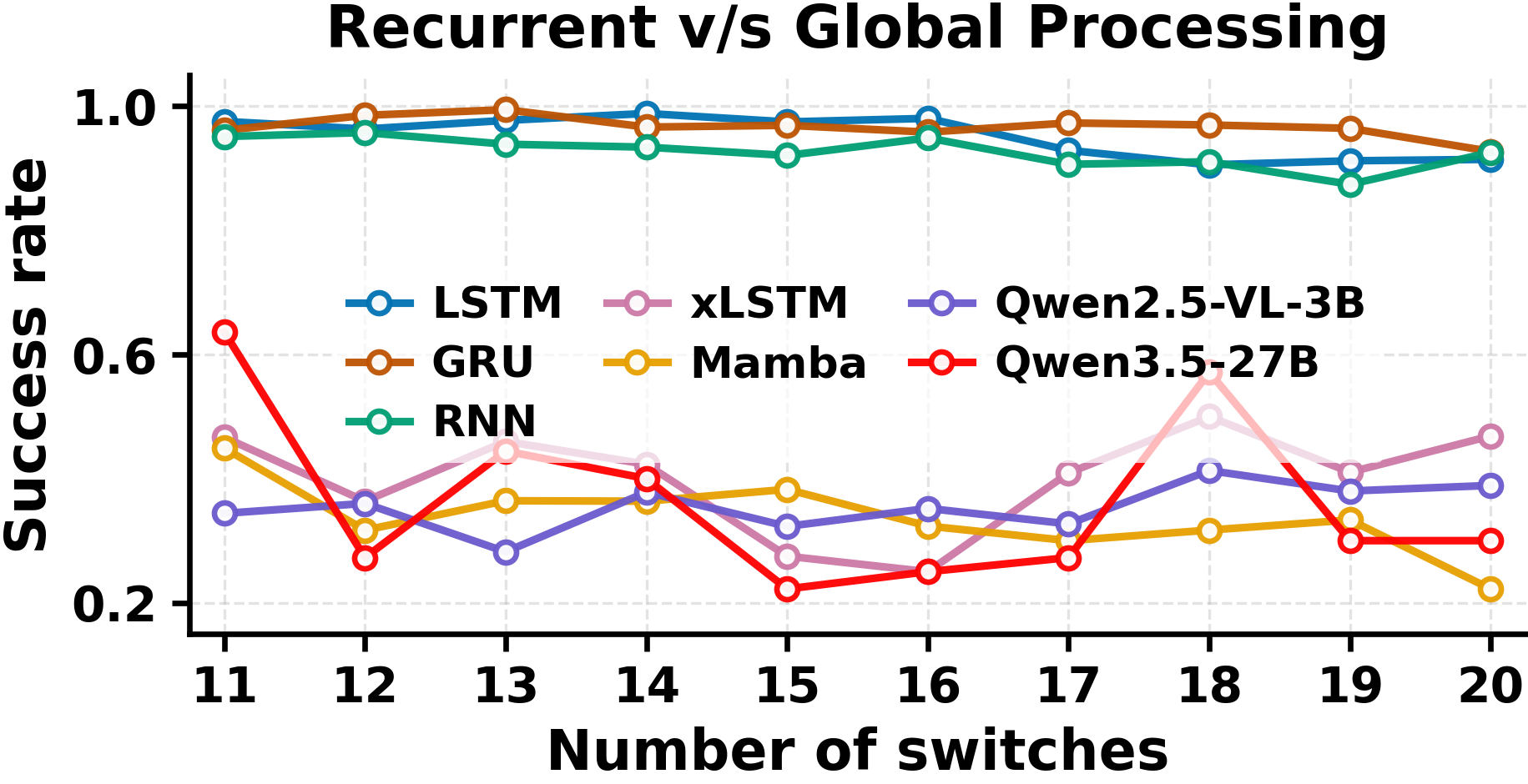}
    \end{minipage}
      \captionof{figure}{
  Effect of sequence-processing architecture on \taskStateMachine{} with fixed local visual inputs.
  Strictly recurrent models generalize out-of-distribution, while all other tested architectures do not.
      }
    \label{fig:add_baselines_sm}
    \vspace{-0.5em}

\end{minipage}

\end{figure}

\subsection{State tracking with active visual search}
\label{parity_active_search}


The state-tracking tasks in \cref{sec:exp_2d_tasks} above use visual cues in the form of arrows to specify the order in which
  the switches should be visited.
  We next study a complementary and more
  challenging variant in which these visual cues are removed. Here, the model
  must
  combine state tracking with active visual search: it must locate an unvisited switch, navigate to it, process its symbol, update the current state, and
  repeat this process until all switches have been considered.

 In this experiment, to make exploration more efficient, 
 we augment the model with zoom-in and 
 zoom-out actions, allowing it to adjust
  peripheral coverage while searching for the next target. 
  As previously, the peripheral glimpse is always down-sampled to a fixed 
  low resolution, regardless of its spatial extent. 

  To further support navigation, we also make it possible for 
  the model to mark each switch it has seen with a black dot on
  the canvas, and all subsequent glimpses reflect this change. 
  The marks serve as an external memory for the model's visitation history. 
  We find that this makes it much easier to train the model. 
  The agent still has to choose its next fixation, interpret each switch locally, and update the current state.

  As shown in \cref{fig:parity_active_search}, with these modifications 
  the model continues to generalize out-of-distribution in this 
  more challenging setting. These results indicate that the 
  generalization behavior observed in the previous 
  task with arrows is not merely a consequence of following an explicitly
  provided traversal path. Rather, when equipped with local recurrent
  processing, low-resolution peripheral search, and a spatial 
  memory of past visitations, the model learns to combine active 
  exploration with state tracking.

  \subsection{Recurrent vs global processing of local visual inputs}


As we discussed, local perception controls how visual information is 
input to the model, but out-of-distribution generalization also depends 
on the computational backbone used to process that information. 
In this experiment, we fix a stream of input glimpses for the 
\taskStateMachine{} task, and we compare the performance of multiple 
different architectures on the same local observations. 
We compare non-linear recurrent networks with models known to 
struggle with length-generalization, including transformers and linear RNNs 
(Mamba and XLSTM). 

\cref{fig:add_baselines_sm} shows that recurrent networks are the 
only models that length-generalize. 
The results are in line with similar findings in textual state tracking tasks,
and they suggest that the non-recurrent models can learn shortcuts 
even when the visual input is local.
We also note that, even though strong pre-trained VLMs such as \qwen{} and QWEN3.5-VL-27B, perform well when fine-tuned in-distribution, 
they do not generalize out-of-distribution. 
Additional \taskVisualParity{} results in
  \cref{appsec:visual_parity_baselines} follow the same trend.

This experiment provides clear evidence in support of \textbf{H2}.

  \subsection{Recall task}

    \begin{wrapfigure}{r}{0.55\linewidth}
    \centering
    \includegraphics[width=\linewidth]{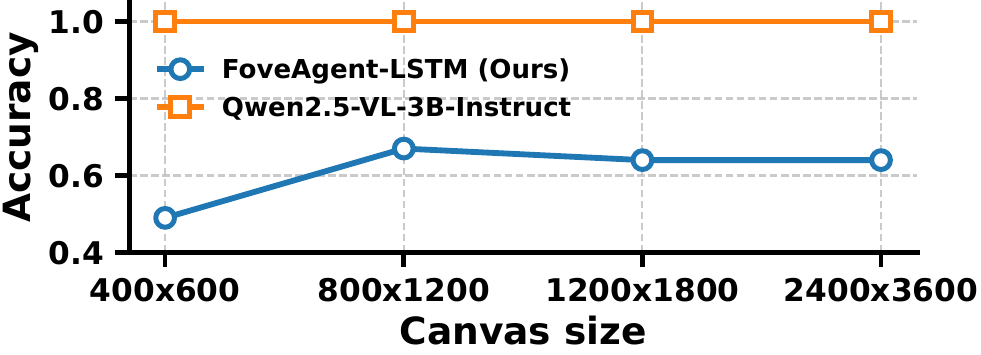}
    \vspace{-2em}
    \caption{Results on \taskRecall{}, a visual retrieval task with increasing clutter but no sequential state tracking.
  The (non-foveated) global model performs well in this setting, highlighting the distinction between recall-heavy visual search and state-tracking tasks.}
  \label{fig:recall}
    
        \vspace{-1.5em}
  \end{wrapfigure}

We now turn to \taskRecall{} to study OOD generalization
in the absence of any state tracking.
We train \model{} and \qwen{} on the task with  results shown in 
\cref{fig:recall}.
We note that \qwen{} significantly outperforms the foveated model 
on this task. 
As we note in \cref{sec:tasks}, this task does not require state-tracking and a global vision model can learn the correct set of binding features \citep{campbell2024understanding}
leading to high accuracy. 
This is true despite using a significantly higher number of 
distractors during testing than during training. 
The fact that the recurrent model generalizes state tracking
but not recall, while the transformer-based model does the converse,
mirrors similar findings in the case of textual tasks \citep{phan2025delayed}.

\subsection{Towards length Generalization in a Real-world Scenario}
Vision-language models (\eg, \citealp{Qwen2.5-VL}) have made
significant progress in reasoning over mathematical plots as
shown through evaluations on benchmarks
like MathVista \citep{lu2024mathvista}.
Here we consider the real-world task of finding roots, 
which should arguably benefit from state tracking capabilities as discussed
in \cref{sec:tasks}. We demonstrate that despite impressive results on standard benchmarks, model performance lags significantly on this task particularly in terms of OOD length generalization.


\begin{table*}[!t]
  \centering
    \caption{
  Accuracy (\%) on \taskFindingRoots{} across in-distribution and OOD test settings; higher is better.
  Rows indicate the evaluation split: OOD-subplots increases the number of subplots, OOD-numroots increases the number of function roots, and OOD-(subplots+nr) increases both.
  The first column is the finetuned global-view \qwen{} baseline using a single global image input, while the remaining columns are \foveaqwen{} variants that combine a global glimpse (G) with high-resolution
  local glimpses (L).
  }
  \label{tab:qwen_plottwist}
  \begin{minipage}{\textwidth}
  \centering
  \begin{tabular}{lcccc}
  \toprule
  {Test Scenario} & {G(1200,800)} & {G(480,320) + L} & {G(600,400) + L} &
  {G(1200,800) + L}\\
  \midrule
  In-distribution & 57.24 & 74.50 & 80.44 & 82.26 \\
  OOD-subplots & 50.12 & 57.49 & 68.78 & 77.24 \\
  OOD-numroots (nr) & 32.63 & 61.62 & 65.85 & 67.12 \\
  OOD-(subplots+nr) & 35.53 & 50.92 & 59.70 & 57.46 \\
  \bottomrule
  \end{tabular}
  \label{tab:qwen_overall}
  \end{minipage}
  \end{table*}
We finetune \qwen{} as well as a ``foveated'' version of \qwen{}, \ie, \foveaqwen{}.
We use \foveaqwen{} instead of \model{} here, as
currently, there are no large pre-trained vision-language models with a recurrent backbone capable of reasoning effectively over mathematical plots.
\foveaqwen{} receives a low-resolution glimpse of the original image, and local foveated glimpse(s) of size $200 \times200$ pixels at the original image resolution.
Both models are trained with \textit{exactly the same} chain-of-thought trajectory, visualized in \cref{fig:full_cot}; however, \qwen{} does not receive local crops at the predicted locations because it already
  has access to the full-resolution global image of the canvas.
During training, the model sees up to 6 subplots on the canvas and functions containing up to 6 roots. On each subplot a maximum of 4 functions may be present.
As OOD scenarios, we consider generalization over the number of subplots [7-9],
the number of zeros [7-10], and a combination of both.



 As shown in \cref{tab:qwen_plottwist}, the foveated approach significantly outperforms the model using only a $1200\times800$ global view, both in InD and OOD settings. These gains suggest 
 that many failures of
  the global model stem from insufficient access to fine-grained 
  visual evidence: 
  local glimpses allow \foveaqwen{} to inspect the relevant 
  subplot, function, and root locations. At the same time, one should 
  interpret the OOD-numroots gains cautiously. 
  The model is evaluated beyond the training range of roots, but this split represents a moderate extrapolation from nearby counts rather 
  than the stronger form of
  systematic length generalization tested in the synthetic state-tracking tasks. 
  Thus, the result supports the importance of foveated perception for 
  gathering visual evidence from the canvas. But they remain consistent 
  with  our broader conclusion that robust OOD state tracking 
  requires recurrent processing.  

  Finally, we observe that for global models, performance degradation for OOD scenarios involving more subplots is substantially smaller than for scenarios involving more roots.
  This aligns with the earlier recall experiments: locating a target subplot is closer to a visual retrieval problem, whereas finding multiple roots requires maintaining and updating state across a variable-
  length sequence of local observations.

\myparagraph{Limits of uniformly scaling resolution:} 
Another effect of adding local glimpses is that it increases the 
compute available to the model by allocating additional tokens 
to selected regions of the image. 
To ensure that the gains from \foveaqwen{}
  are not simply due to using more visual tokens, 
  we compare against global-view baselines at matched 
  computational budgets. For a fair comparison, each global-view baseline is finetuned for its corresponding input, 
  allowing each baseline to adapt to the corresponding visual token budget.

  \begin{figure*}[t]
      \centering
      \includegraphics[width=0.9\textwidth]{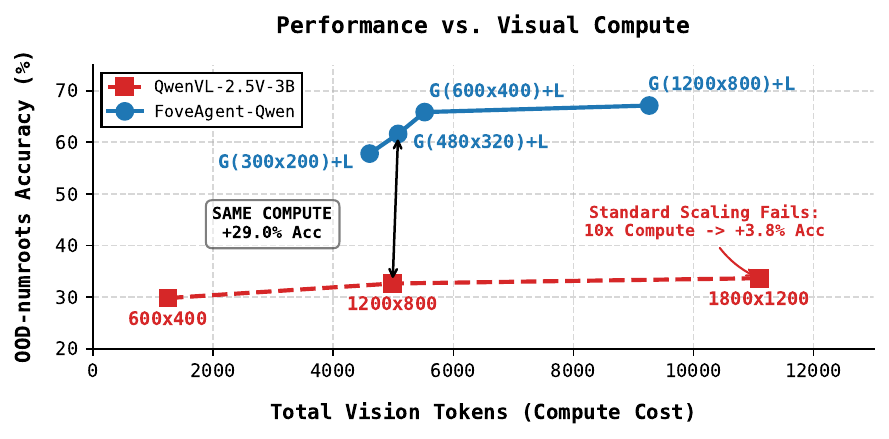}
      \vspace{-0.8em}
      \caption{Compute-matched comparison on \taskFindingRoots{}. For the same visual compute budget, \foveaqwen{} nearly doubles the accuracy of the finetuned global-view \qwen{} baseline, showing that allocating
  compute to high-resolution local glimpses is substantially more effective than uniformly increasing global image resolution.}
      \label{fig:compute_tradeoff}
      \vspace{-1.0em}
  \end{figure*}

\cref{fig:compute_tradeoff} shows that \foveaqwen{} nearly doubles the accuracy of the baseline for the same compute budget. 
This result suggests a clear functional division: global resolution needs 
to be sufficient only for coarse navigation, 
while fine-grained visual reasoning is better served by high-resolution local glimpses. 
Notably, even when the global-view baseline is finetuned at substantially higher resolution, uniformly scaling global compute yields only marginal 
gains, indicating that the advantage of \foveaqwen{} is not simply 
a consequence of using more visual tokens.

\section{Related work}
\myparagraph{OOD Generalization in Language Tasks.}
It is well known that transformer-based language models fail at performing algorithmic tasks
in token sequences whose lengths are different from those seen during training \citep{dubois2020location,anil_lengthgeneralization,faithandfate,zhou2024algorithms}.
Several recent works have proposed explanations for this inability, ranging from
the inability to solve tasks with high ``globality'' \citep{NEURIPS2024_3107e4bd}
to the inability to perform random-access addressing into the context window \citep{ebrahimi2024your}.
Recently, it has been shown that the problem is not only restricted to transformer-based models
but also holds for linear RNNs and state-space models (e.g.,
\citealp{merrill2024illusion, sarrof2024the,cirone2024theoretical,shakerinava2026the,ebrahimi2025revisiting}).
In this work, we consider the related problem of performing algorithmic reasoning
in the visual domain, and we show that there exist important commonalities but also differences
to reasoning over symbol sequences.

\myparagraph{OOD Generalization in Vision Tasks.}
There is a wide range of benchmarks on visual reasoning
(e.g., \citealp{lu2024mathvista,yue2024mmmumassivemultidisciplinemultimodal} and references therein)
as well as a wide range of benchmarks that measure OOD generalization in
vision, such as  \citep{babaiee2025visual,Cherian2023mar,zhao2022oodcvbenchmarkrobustnessoutofdistribution}.
In contrast to the existing work, in this work
we consider a highly specific and simple
form of OOD generalization: the ability
of a model to generalize over a number of visual entities that
differs between training and testing.
This is in direct analogy to existing studies in the language domain \citep{dubois2020location,anil_lengthgeneralization,faithandfate}, where
it was shown that this is type of generalization is challenging for existing foundation
models, although it is trivial for humans.
A simple visual version of the parity task, similar to some of the tasks we
study in the work, is discussed by \citet{stabinger2020training}. In contrast
to our work, that work considers in-distribution performance
(and specifically failures of gradient-based learning),

Similar to our work, \citet{NEURIPS2023_3a40e042} study the use of
recurrent networks for visual OOD generalization.
However, in that work, models perceive whole images whereas in our study
they perceive local regions which improves OOD generalization.
Importantly, the results in that work show that, although recurrent models outperform
non-recurrent ones, generalization performance drops sharply as a function
of image size, in line with our findings in \cref{section:roleoflocalperception}.
Similarly, \citet{lotfi2025chainofsketchenablingglobalvisual} present an approach to OOD
generalization by training a convolutional network to perform (global)
pre-defined modifications to the input image.
In contrast to that work, we investigate the impact of local versus global
perception on the ability to generalize and show that local perception is an inductive
bias that enables generalization.

\myparagraph{Visual attention.} Foveation is an important feature of human vision, and there is a range of
work exploring the use of biologically inspired spatial attention in computer vision tasks, including
\citep{NIPS2010_677e0972,ba2014multiple,mnih2014recurrent,Gregor2015DRAWAN}
(see also \citealp{visualattentionsurvey} for a recent survey).
However the study of spatial attention has taken a back-seat in recent years
in favor vision-language models with convolutional or ViT-based visual front-end.
In contrast to existing work in that area, our primary goal is not to introduce a novel attention mechanism but to study the way in which local attention can benefit length-generalization.

\section{Discussion}
In this work, we study out-of-distribution generalization in the context of visual reasoning. 
We hypothesize that the common practice of encoding visual information 
in a global context window, or by using a global vision front-end, such as 
a convolutional network or ViT, prevents existing models from generalizing 
out-of-distribution. 
We show that combining local visual input with recurrent processing 
unlocks length generalization in visual state tracking.
These results suggest recurrence and locality both to be necessary,
and the combination of both to be sufficient for compositional generalization 
in visual reasoning tasks.

We also show that out-of-distribution generalization on the one hand,
and the need to explore or perform search over the image on the other, 
are in conflict, in that the first requires 
small/low-resolution glimpses while the latter requires large/high-resolution 
glimpses. 
And we show that a human-inspired foveation model combining a 
high-resolution fovea with a low-resolution periphery makes it possible  
to resolve this conflict. 

We train the local perception policies in this work using imitation learning based on
ground truth policies. While these are straightforward to generate for synthetic tasks,
learning broadly general policies is an open research problem and it 
will likely require more varied tasks, reinforcement learning or 
a combination of both.

Recent work has shown that recurrent computation in language models 
not only improves OOD generalization, but that it also has a dramatic impact
on data efficiency for learning reasoning tasks \emph{in distribution} \citep{ebrahimi2026on}.
We hypothesize that a similar finding will be true for visual reasoning 
tasks, which would suggest that local visual attention may be a crucial, 
albeit vastly underexplored, ingredient in future visual foundation models.

\bibliography{main}
\bibliographystyle{iclr2026_conference}

\appendix
\newpage
\appendix

  \section{Appendix Overview}
  \label{appsec:overview}

  The Appendix is organized as follows:
  \begin{enumerate}
      \item \cref{appsec:evaluation_metrics} describes the evaluation metrics used for each task.
      \item \cref{appsec:dataset_details} provides dataset and task-construction details, including the formal construction of the \taskStateMachine{} task.
      \item \cref{appsec:implementation_details} gives implementation and training details for \model{} and \foveaqwen{}.
      \item \cref{appsec:controlled_results} reports additional controlled visual-reasoning results that support the main-paper analyses of local versus global perception and recurrent versus global processing.
      \item \cref{appsec:finding_roots_results} provides additional \taskFindingRoots{} results and qualitative examples, including the multi-function split and visual trajectory generation procedure.
      \item \cref{appsec:zeroshot_prompts} lists the zero-shot VLM prompts for \taskVisualParity{}, \taskStateMachine{}, and \taskRecall{}.
  \end{enumerate}

\section{Evaluation Metrics}
\label{appsec:evaluation_metrics}

In this section, we describe the criteria used to evaluate the models on all the tasks considered in this work.
 Evaluation for \textit{Visual-Parity} and \textit{State-Machine} is based on prediction accuracy, requiring the model to correctly identify the final state (e.g., parity 0/1 or state machine 0/1/2). Crucially, for \model{}, we impose a traversal requirement: the model must attend to every task-relevant object on the canvas. Consequently, any instance where \model{} misses an item is penalized as incorrect, even if the classification is coincidentally correct.
For \textit{Conjunctive Search} task, we also use the accuracy metric.
For the \textit{Finding Zeros} task, we implement a precision-based evaluation metric that accommodates reasonable approximation in numerical predictions. Using the metadata available in Math-Search~\citep{madhan2024mathsearch}, we conduct a relaxed evaluation for each predicted zero. A prediction is considered correct if it falls within one quarter of one tick step of the x-axis of the ground truth value. For example, in \cref{fig:plottwist_example}, where the x-tick step is $1.0$, the predicted root must be within $\pm 0.25$ units of the actual root to be counted as correct. Since a function may have multiple zeros, we use partial scoring i.e. the model is assigned a score proportional to the number of zeros predicted correctly. Concretely, the score takes the form $S(x) = \frac{\sum_i \mathbf{I}(x_i \sim\in Z)}{\left| x \right|} $ where $x$ is the set of zeros predicted by the model, $Z$ is the set of ground truth zeros and $\sim\in$ represents the relaxed evaluation as described above.

 \section{Dataset and Task Details}
  \label{appsec:dataset_details}

  In this section, we provide additional details about the datasets and task constructions used in the main paper.
  For each task described in Sec.~3.1, we generate in-distribution (InD) and out-of-distribution (OOD) splits by varying image resolution and task complexity, as summarized in Sec.~3.2.
  All models are trained only on InD training examples and evaluated on both InD and OOD test sets.
  For each task, we generate separate training, validation, and InD test sets, as well as additional OOD test sets.
  For \model{}, training examples are generated using oracle policies that produce trajectories of glimpse-action pairs.
  Each trajectory contains local glimpses, peripheral glimpses, actions, probes, the task question, and the final answer label.
  For \taskStateMachine{} and \taskVisualParity{}, the switches on canvas are connected with arrows. The oracle policy follows the arrows until the last unvisited switch. 
  The step size of the oracle policy is bounded by the size of the peripheral glimpse, $G_p$. More concretely, it takes a step of up to a maximum of $0.8\times |G_p|$ in the direction of the next unvisited switch.
  For the more challenging variant of \taskVisualParity{} described in \cref{parity_active_search}, the switches are not connected with arrows. Here, the oracle policy visits the next unvisited switch visible in the peripheral glimpse. 
  If no unvisited switch is visible, the policy zooms out until it finds an unvisited switch, or until 
  all the boundaries of the canvas are visible. When all boundaries of the canvas are seen the episode is ended.  
  Each zoom-out action doubles the size of glimpses i.e. the glimpse sees 4x the area compared to the previous step. Once the agent reaches the target, the policy zooms in to bring both the glimpse sizes to the original size. Similar to zoom-out, each zoom-in action halves the size of the glimpses.
  
  It is to be noted that the regardless of the size, the glimpses are always down-samp
  led to the sensor resolution before they are passed as inputs to the model to avoid learning shortcuts.
  For robustness, we add noise to the positions of the glimpses in the oracle trajectories so that the trained policy can learn recovery behavior.
  For all tasks except \taskFindingRoots{}, we generate 50k training trajectories at a resolution of $1200\times800$.
  Evaluation is conducted on 2000 unseen samples from each InD and OOD split.

\begin{figure*}[!ht]
    \centering
    \includegraphics[width=0.95\linewidth]{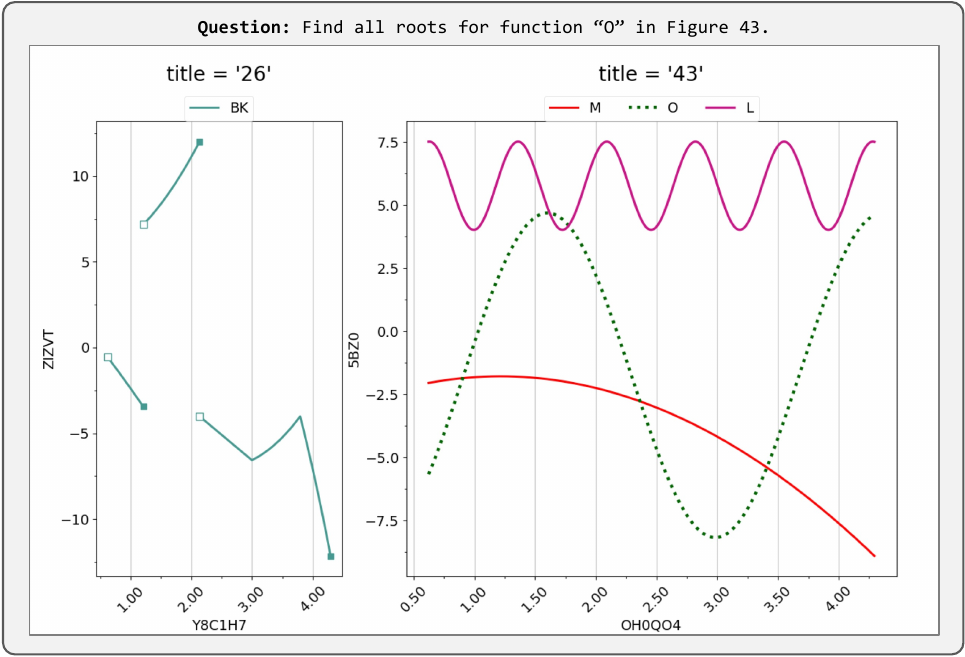}
    \caption{Example from \taskFindingRoots{}. The model must identify the target subplot and function from the question, then estimate the x-values where the target function crosses zero.}
    \label{fig:plottwist_example}
\end{figure*}

\begin{figure*}[!ht]
    \centering
    \includegraphics[width=0.99\linewidth]{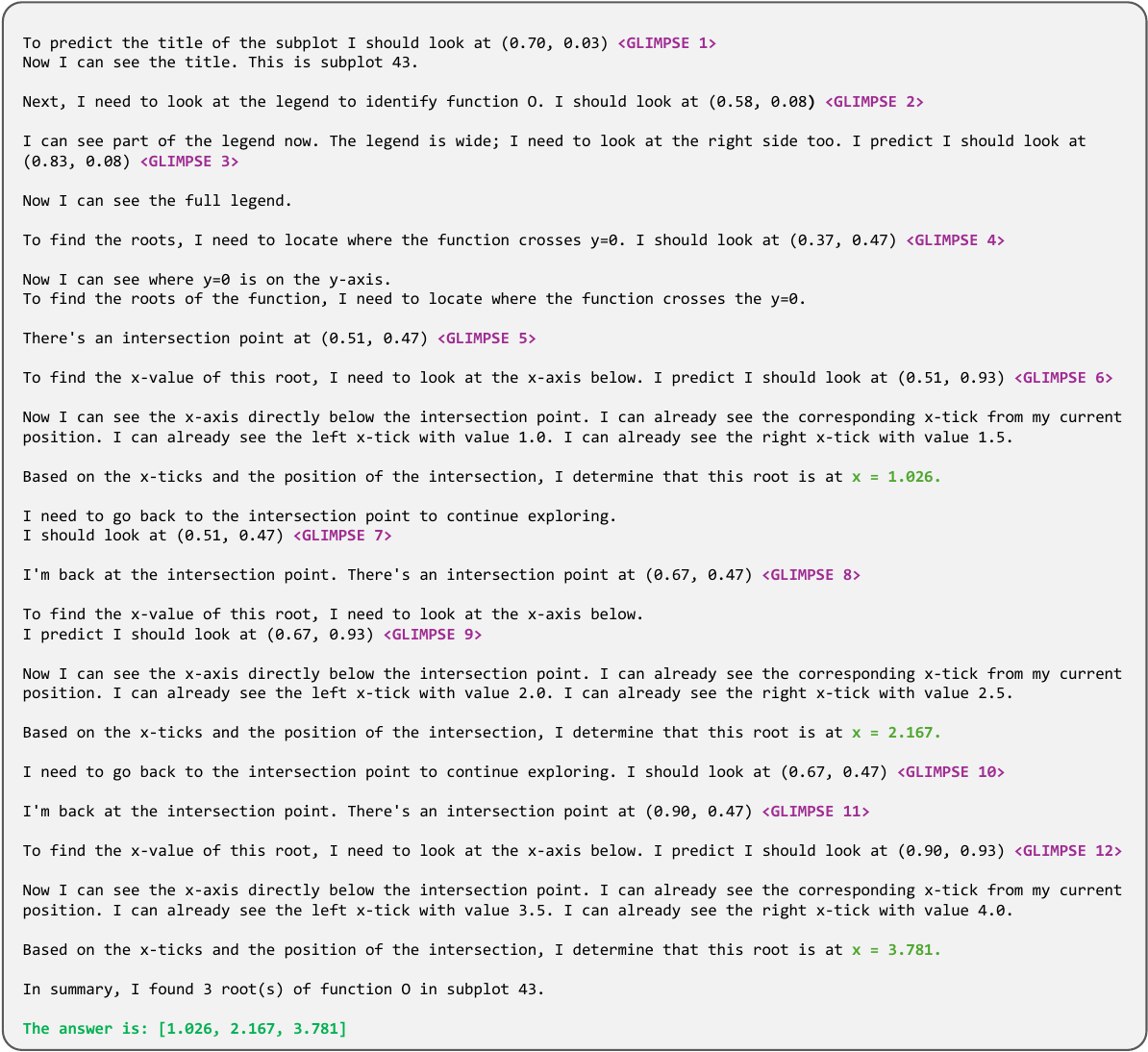}
    \caption{Visual chain-of-thought trajectory for the \taskFindingRoots{} example in \cref{fig:plottwist_example}. The predicted fixation locations determine the high-resolution local crops supplied to \foveaqwen{}, enabling fine-grained inspection of the target subplot and root locations. The finetuned global-view \qwen{} baseline is trained with the same chain-of-thought trajectory, but does not receive these local crops because it already has access to the full-resolution global image.}
    \label{fig:full_cot}
\end{figure*}

  For \taskFindingRoots{}, we curate a training set of 20k samples from Math-Search~\citep{madhan2024mathsearch}, containing images of size $1200\times800$, question-answer pairs, and rich metadata for each plot.
  For each sample, we generate a visual trajectory that simulates an agent equipped with both global and local glimpses as it locates the roots of a target function in a specified subplot.
  The agent begins at the upper-left corner $(0,0)$ and navigates to positions that help identify the correct subplot and target function before executing the root-finding procedure.
  The complete pseudocode for trajectory generation is provided in \cref{alg}.
  An example of this task is shown in \cref{fig:plottwist_example}, and the corresponding trajectory generated by \foveaqwen{} is shown in \cref{fig:full_cot}.

  \myparagraph{\taskStateMachine{} construction.}
  The main text introduces \taskStateMachine{} as an order-sensitive visual reasoning task: unlike \taskVisualParity{}, it is not enough to aggregate the set of observed switch values; the model must process them
  in the order specified by arrows.
  Here we give the formal construction underlying that task.
  Following \citet{liu2023transformers} (Example 6), we instantiate the visual sequence as a state machine realizing the action of the dihedral group.
  Each switch corresponds to one input symbol, and the arrows in the image specify the order in which those symbols must be processed.
  This construction is useful because it creates a simple state-tracking problem where the correct answer depends on sequential updates, making shortcut solutions based on unordered aggregation insufficient.

  Formally, we consider a semiautomaton with state space
  $Q=\{0,\ldots,S-1\}\times\{-1,+1\}$
  and input alphabet
  $\Sigma=\{\texttt{advance},\texttt{reverse}\}$.
  The transition function $\delta: Q\times \Sigma \to Q$ is defined as
  \begin{align*}
  \delta((s,b),\texttt{advance})
      &= (s+b \!\!\!\pmod S,\; b), \\
  \delta((s,b),\texttt{reverse})
      &= (s,\,-b),
  \end{align*}
  where $s$ denotes the current position on a cycle of length $S$, and
  $b\in\{-1,+1\}$ specifies the current direction.
  Thus, \texttt{advance} moves one step in the current direction, while
  \texttt{reverse} flips the direction of subsequent movements.
  The transformations induced by these operations realize the action of the dihedral group $D_S$.
  In all experiments, we set $S=3$.

  Given an input sequence $\sigma_1,\ldots,\sigma_n \in \Sigma^n$, represented visually by the sequence of switches connected by arrows, the state evolves according to
  \[
  q_i = \delta(q_{i-1}, \sigma_i), \qquad i=1,\ldots,n,
  \]
  starting from the initial state $q_0=(s_0,b_0)$ specified in the input prompt.
  Let $q_n=(s_n,b_n)$ denote the final state after processing the entire input sequence.
  The task is to determine the final position $s_n \in \{0,1,2\}$.
\section{Implementation Details}
  \label{appsec:implementation_details}

In this section, we describe the implementation details of \model{}. We use a 4-layer LSTM~\citep{hochreiter1997long} as our base recurrent model and a ResNet-18~\citep{he2016deep} as our visual backbone. \model{} takes 2 glimpses as input at each step. A local glimpse of resolution $80 \times 80$ and a peripheral glimpse that has a resolution of $x-\text{multiple}$ where $x \in \{1, 2, 4, 8, 16\}$. We use $x=4$ unless otherwise specified. Refer to Sec 5.4
for a sweep over $x$. At each step, we have three 2-layer MLPs that predict \textit{probes} (what the current glimpse is looking at), \textit{actions} (where to look at next) and \textit{task} (the final output of the task, which is only supervised for the last step) respectively. We train all our models using Adam optimizer with a learning rate of $1\times10^{-4}$ on 4 Nvidia A100 GPUs with a batch size of 48 (12 on each GPU). We train \model{} for 200k iterations and evaluate every 10k iterations on a val set. We select the best performing checkpoint on the val set, and report its performance on a test set. \\

To develop \foveaqwen{}, we utilize \qwen{} as our foundation model. As a baseline, we fine-tune the base model on the \textit{Finding Zeros} task. Following the procedure in \cref{alg}, the baseline \qwen{} model receives a full-resolution global image ($1200 \times 800$) and simulates an \textit{imaginary} glimpse of size $200\times200$ to locate function roots. \\
In contrast, our \foveaqwen{} follows an identical trajectory but supplements the global glimpse with local glimpses corresponding to predicted locations, enabling fine-grained visual processing of complex subplots. For both configurations, we fine-tune the vision encoder and language modeling backbone of \qwen{} on 20,000 samples over 10 epochs, with a batch size of 4 distributed across 4 Nvidia A100 GPUs (1 sample per GPU) using the Adam optimizer with a learning rate of $1\times10^{-5}$. Similar to \model{}, we select the best-performing model based on validation performance and report its results on InD test sets and various OOD scenarios in Table 1.
In the table, the first column presents results from the baseline model (finetuned with full high-resolution global image), while subsequent columns show \foveaqwen{} performance with varying lower global glimpse resolutions but a fixed local glimpse size of
$200\times200$.

\section{Additional Controlled Visual Reasoning Results}
  \label{appsec:controlled_results}

  The main paper separates two ingredients required for OOD visual reasoning:
  \textit{how visual evidence is gathered} and \textit{how the resulting observations are processed}.
  Section~\ref{sec:exp_2d_tasks} shows that \model{} generalizes on the 2D state-tracking tasks, while the following sections isolate these two factors more directly.
  In particular, the \textit{Local vs Global Perception} experiment studies the visual interface, and the \textit{Recurrent vs Global Processing of Local Visual Inputs} experiment studies the computational
  structure used after local observations have been obtained.
  Here, we provide additional controlled results that support both conclusions.

  \subsection{1D Visual Parity: Controlling for the Visual Front-End}

  \begin{figure}[t]
      \centering
      \includegraphics[width=0.6\textwidth,trim={0cm 0 0cm 0},clip]{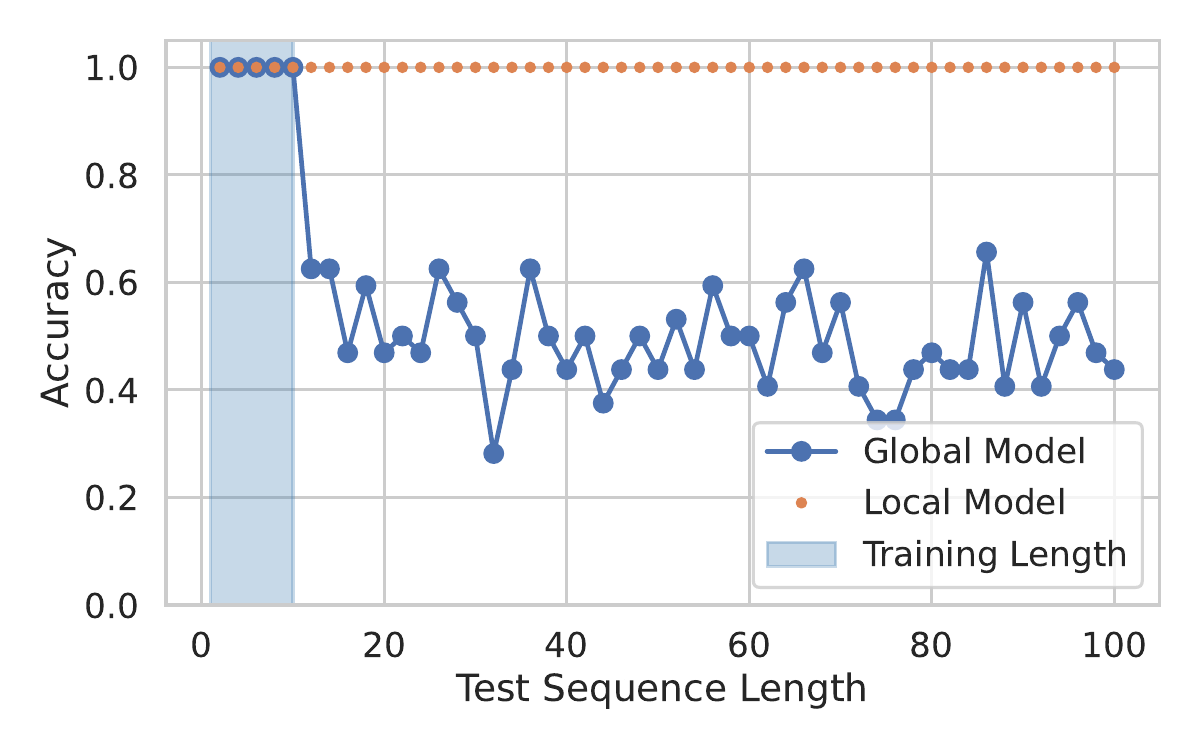}
      \caption{Results on the 1D visual parity task. The local recurrent model generalizes to longer sequences than seen during training, while the global model fails outside the training-length regime.}
      \label{fig:oned_results}
  \end{figure}

  The 2D visual parity task differs from the standard language parity task in two ways:
  the input is spatially distributed over an image, and the model must use a visual front-end to extract the relevant symbols.
  To control for this difference, we construct a simpler 1D visual parity task that keeps the input visual while making the sequence structure explicit.
  We render the bit sequence as a horizontal image, with white and gray blocks denoting 0s and 1s, respectively, and a black block marking the end of the sequence.

  This setting connects directly to the main-paper discussion on processing local visual inputs.
  Once the relevant symbols are presented in a simple sequential format, the central question becomes whether the model can apply the same parity update repeatedly beyond the training length.
  As shown in \cref{fig:oned_results}, the local recurrent model generalizes to substantially longer sequences than those seen during training.
  The global model, in contrast, achieves strong in-distribution performance but fails to generalize to longer OOD sequences.
  This mirrors the gap between recurrent networks and transformer-style models on language parity, and supports the main-paper conclusion that recurrence is important for robust length generalization.

  \subsection{Additional Visual Parity Baselines}
  \label{appsec:visual_parity_baselines}

  In the main paper, the \textit{Local vs Global Perception} experiment shows that providing a recurrent model with the full high-resolution canvas can encourage shortcut learning.
  Specifically, the \textit{Global} and \textit{Local+Global} variants perform well in distribution but degrade OOD, whereas \model{} maintains high accuracy by combining local high-resolution glimpses with low-
  resolution peripheral context.
  This indicates that local perception is not merely an implementation detail; it is an important inductive bias for preventing global shortcuts.

  The \textit{Recurrent vs Global Processing of Local Visual Inputs} experiment then fixes the visual input stream and varies only the architecture used to process it.
  That comparison shows that strict recurrent state updates generalize OOD, while architectures with richer global interactions over the observation history, including attention-based transformers, Mamba, and
  xLSTM, degrade more sharply.
  Together, these two main-paper experiments suggest that OOD visual state tracking requires both an appropriate local visual interface and a computation that scales through repeated recurrent updates.

  Here we provide an additional comparison on \taskVisualParity{} that reinforces the same conclusion.
  We evaluate more recurrent backbones and a VLM baseline on increasing task lengths.
  As in the main-paper \taskStateMachine{} results, models that strictly process observations step by step maintain stronger OOD generalization.
  By contrast, backbones that permit non-strictly-recurrent or global processing can fit the training-length regime but are more prone to shortcut solutions that fail as the number of switches increases.

  \begin{figure}[ht]
      \centering
      \includegraphics[width=0.75\textwidth]{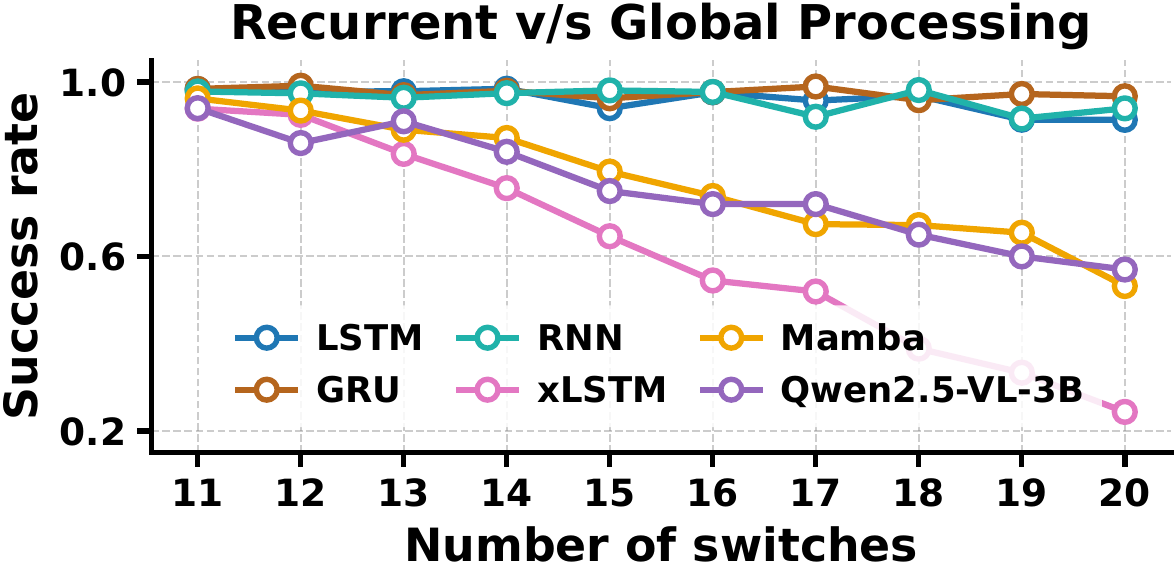}
      \caption{Additional baselines on \taskVisualParity{} under increasing task length. Strict recurrent backbones maintain stronger OOD generalization, while non-recurrent or globally processing baselines
  degrade as the number of switches increases.}
      \label{fig:add_baselines}
  \end{figure}

\FloatBarrier
\section{Additional Results on \taskFindingRoots{}}
  \label{appsec:finding_roots_results}
  
In this section, we report results on a more challenging subset of \taskFindingRoots{} where the target subplot contains two or more functions. This setting requires finer visual processing to identify the target function and estimate its roots, making the benefit of local glimpses especially clear.

  \begin{table*}[t]
  \centering
  \caption{Accuracy (\%) on the multi-function split of \taskFindingRoots{}; higher is better. Rows indicate InD and OOD test settings, where OOD-subplots increases the number of subplots, OOD-numroots increases the number of roots, and OOD-(subplots+nr) increases both. The global-view \qwen{} baseline receives only a full-image input G, while \foveaqwen{} variants combine a global glimpse G with high-resolution local glimpses L.}
  \label{tab:qwen_mf}
  \small
  \begin{tabular}{lccccc}
  \toprule
  Test Scenario & G(1200,800) & G(300,200)+L & G(480,320)+L & G(600,400)+L & G(1200,800)+L \\
  \midrule
  In-distribution       & 26.63 & 41.52 & 60.59 & 65.19 & 67.01 \\
  OOD-subplots          & 19.75 & 33.10 & 36.52 & 49.90 & 65.13 \\
  OOD-numroots (nr)     & 1.06  & 47.75 & 56.30 & 62.37 & 63.66 \\
  OOD-(subplots+nr)     & 7.90  & 35.08 & 35.86 & 52.07 & 45.96 \\
  \bottomrule
  \end{tabular}
  \end{table*}

\begin{algorithm}
\caption{Oracle visual trajectory for \taskFindingRoots{}. The trajectory first localizes the target subplot and legend, then scans along the zero level to identify roots and interpolate their x-values when needed.}
\begin{algorithmic}[1]
\State Initialize at top-left corner $(0, 0)$
\State Navigate to the target subplot title
\State Check if the legend box is fully visible
\If{legend not fully visible}
    \State Traverse the legend box to explore all entries
\EndIf
\State Locate or interpolate $y=0$ on the y-axis
\State Begin horizontal scan from y-axis
\While{not reached end of function plot}
    \State Stop at intersection where $y=0$ meets the function
    \If{corresponding x-value is visible}
        \State Continue search for next intersection point
    \Else
        \State Traverse downward to the x-axis
        \If{surrounding x-ticks not visible}
            \State Traverse horizontally to locate surrounding x-tick values
            \State Calculate the target x-value through interpolation
        \EndIf
        \State Return to the intersection point
    \EndIf
\EndWhile
\label{alg}
\end{algorithmic}
\end{algorithm}

\clearpage
\section{Zero-Shot Evaluation Prompts}
  \label{appsec:zeroshot_prompts}

\newcommand{\prompttag}[1]{\texttt{\footnotesize\textless #1\textgreater}}
\newcommand{\promptendtag}[1]{\texttt{\footnotesize\textless/#1\textgreater}}

The following prompts are used for zero-shot VLM evaluations. We typeset them directly for readability; the required response format is always to reason inside \prompttag{think} tags and provide the final answer inside \prompttag{ans} tags.

\subsection{\taskVisualParity{} Prompt}
\label{fig:zs_parity}
\begin{tcolorbox}[breakable, enhanced, colback=gray!4, colframe=black!45, coltitle=black, title={System prompt for Visual Parity}, fonttitle=\bfseries, boxrule=0.4pt, arc=1mm, left=5pt, right=5pt, top=5pt, bottom=5pt]
You are a helpful visual assistant that analyzes images of nodes to determine the parity of the bit string rendered over the image.

The image contains a number of circular nodes (blobs). Each blob has a digit \texttt{0} or \texttt{1} written in its center. You must find all the blobs in the image, extract the bit string, and compute its parity.

\textbf{First, understand how to classify each node:}
\begin{itemize}[leftmargin=*]
    \item Value 1: The node has the digit \texttt{1} written in its center.
    \item Value 0: The node has the digit \texttt{0} written in its center.
\end{itemize}

\textbf{When analyzing the image to find the parity of the bit string:}
\begin{enumerate}[leftmargin=*]
    \item Scan the image to locate all the circular nodes.
    \item For each node, read the digit (\texttt{0} or \texttt{1}) at its center.
    \item Collect all the identified bits to form a single string. The order of bits does not matter for parity calculation.
    \item Compute the parity of the final bit string. Parity is 1 if there is an odd number of 1s, and 0 if there is an even number of 1s.
    \item Return your final parity value.
\end{enumerate}

For each question, within \prompttag{think} tags, describe each node you find and identify its bit value (0 or 1). Finally, state the extracted string, compute the parity, and provide your final answer as a 0 or 1 inside \prompttag{ans} tags.
\end{tcolorbox}

\clearpage
\subsection{\taskStateMachine{} Prompt}
\label{fig:zs_state_machine}
\begin{tcolorbox}[breakable, enhanced, colback=gray!4, colframe=black!45, coltitle=black, title={System prompt for State Machine}, fonttitle=\bfseries, boxrule=0.4pt, arc=1mm, left=5pt, right=5pt, top=5pt, bottom=5pt]
You are a helpful visual assistant that analyzes images of a graph to determine the final state of a state machine.

You will be provided with an initial state (0, 1, or 2) in the question. The image contains a number of circular nodes (blobs) connected by arrows. Each blob has a digit \texttt{0} or \texttt{1} written in its center. You must follow the arrows to read the sequence of bits and compute the final state.

\textbf{First, understand how to identify the sequence of the nodes based on their colors:}
\begin{itemize}[leftmargin=*]
    \item Green node: The starting node.
    \item Beige nodes: The intermediate nodes.
    \item Purple node: The ending node.
\end{itemize}

\textbf{Next, understand the rules of the state machine:}
\begin{itemize}[leftmargin=*]
    \item You start with the initial state provided in the question.
    \item At the beginning, the operation associated with the symbol \texttt{0} is ``increment''.
    \item Value \texttt{0}: Apply the current operation associated with \texttt{0}. If it is incrementing, add 1 to the state (modulo 3). If it is decrementing, subtract 1 from the state (modulo 3).
    \item Value \texttt{1}: Reverse the operation associated with \texttt{0}. If \texttt{0} meant increment, it now means decrement. If \texttt{0} meant decrement, it now means increment. Encountering a \texttt{1} does not change the numerical state itself.
    \item All state updates are modulo 3, meaning the numerical state is always 0, 1, or 2.
\end{itemize}

\textbf{When analyzing the image to find the final state:}
\begin{enumerate}[leftmargin=*]
    \item Note the initial state given in the question.
    \item Scan the image to locate the green starting node.
    \item Follow the arrows sequentially from the green node, through all beige intermediate nodes, until you reach the purple ending node.
    \item For every node in this sequence (starting with the green node), read its digit (\texttt{0} or \texttt{1}) and apply the rules above to update either the numerical state or the meaning of \texttt{0}.
    \item Return your final state value after processing the entire sequence, up to and including the purple node.
\end{enumerate}

For each question, within \prompttag{think} tags, state the initial state. Then, describe the sequence of nodes you follow. For each node, state its color, identify its bit value (0 or 1), and explain the updated state and the current meaning of \texttt{0} (increment or decrement). Finally, provide your final computed state value as a 0, 1, or 2 inside \prompttag{ans} tags.
\end{tcolorbox}

\clearpage
\subsection{\taskRecall{} Prompt}
\label{fig:zs_conjunctive}
\begin{tcolorbox}[breakable, enhanced, colback=gray!4, colframe=black!45, coltitle=black, title={System prompt for Recall}, fonttitle=\bfseries, boxrule=0.4pt, arc=1mm, left=5pt, right=5pt, top=5pt, bottom=5pt]
You are a helpful visual assistant that analyzes images and finds if a particular object is present in the image or not.

\textbf{When analyzing an image to find the object:}
\begin{enumerate}[leftmargin=*]
    \item Go through each object step by step and classify.
    \item Identify the correct object on the canvas if present.
    \item Return True if the object is present, else return False.
\end{enumerate}

\textbf{Examples:}
\begin{itemize}[leftmargin=*]
    \item Example 1: Question: On the canvas, is a red L present? Answer: \prompttag{think} \ldots{} \promptendtag{think} \prompttag{ans}True\promptendtag{ans}
    \item Example 2: Question: On the canvas, is a red T present? Answer: \prompttag{think} \ldots{} \promptendtag{think} \prompttag{ans}False\promptendtag{ans}
    \item Example 3: Question: On the canvas, is a green L present? Answer: \prompttag{think} \ldots{} \promptendtag{think} \prompttag{ans}False\promptendtag{ans}
    \item Example 4: Question: On the canvas, is a green T present? Answer: \prompttag{think} \ldots{} \promptendtag{think} \prompttag{ans}True\promptendtag{ans}
\end{itemize}

For each question, within \prompttag{think} tags, analyze the image carefully, reason step by step about what you observe, and then at the end, provide your final answer as True or False inside \prompttag{ans} tags.
\end{tcolorbox}

\end{document}